\documentclass[10pt,journal,compsoc]{IEEEtran}
\usepackage{graphicx}
\usepackage{multirow}
\usepackage{amsmath,amssymb}
\usepackage[noend]{algpseudocode}
\usepackage{algorithmicx,algorithm}
\usepackage[colorlinks,linkcolor=red]{hyperref}

\ifCLASSOPTIONcompsoc
  \usepackage[nocompress]{cite}
\else
  \usepackage{cite}
\fi
\ifCLASSINFOpdf
\else
\fi
\begin{document}
	
\makeatletter
\def\endthebibliography{%
	\def\@noitemerr{\@latex@warning{Empty `thebibliography' environment}}%
	\endlist
}
\makeatother
%
\title{Towards Fine-grained 3D Face Dense Registration: An Optimal  Dividing and Diffusing Method}
%
%
%
%
	
\author{Zhenfeng~Fan, Silong~Peng, and~Shihong~Xia
\IEEEcompsocitemizethanks{\IEEEcompsocthanksitem The corresponding author: Shihong~Xia\protect\\
Institute of Computing Technology, Chinese Academy of Sciences, Beijing, China, 100190\protect\\
E-mail: xsh@ict.ac.cn
}
}
\IEEEtitleabstractindextext{%
\begin{abstract}
Dense vertex-to-vertex correspondence between 3D faces is a fundamental and challenging issue for 3D\&2D face analysis. While the sparse landmarks have anatomically ground-truth correspondence, the dense vertex correspondences on most facial regions are unknown. In this view, the current literatures commonly result in reasonable but diverse solutions, which deviate from the optimum to the 3D face dense registration problem. In this paper, we revisit dense registration by a dimension-degraded problem, \emph{i.e.} proportional segmentation of a line, and employ an iterative dividing and diffusing method to reach the final solution uniquely. This method is then extended to 3D surface by formulating a local registration problem for dividing and a linear least-square problem for diffusing, with constraints on fixed features. On this basis, we further propose a multi-resolution algorithm to accelerate the computational process. The proposed method is linked to a novel local scaling metric, where we illustrate the physical meaning as smooth rearrangement for local cells of 3D facial shapes. Extensive experiments on public datasets demonstrate the effectiveness of the proposed method in various aspects. Generally, the proposed method leads to coherent local registrations and elegant mesh grid routines for fine-grained 3D face dense registrations, which benefits many downstream applications significantly. It can also be applied to dense correspondence for other format of data which are not limited to face. The core code will be publicly available at~\href{https://github.com/NaughtyZZ/3D_face_dense_registration}{https://github.com/NaughtyZZ/3D\_face\_dense\_registration}.
\end{abstract}

}

\maketitle

\IEEEdisplaynontitleabstractindextext

%
\IEEEpeerreviewmaketitle

\IEEEraisesectionheading{\section{Introduction}\label{sec:introduction}}

%
%
%
%
\begin{figure*}[htbp]
	\begin{center}
		\includegraphics[width=1\linewidth]{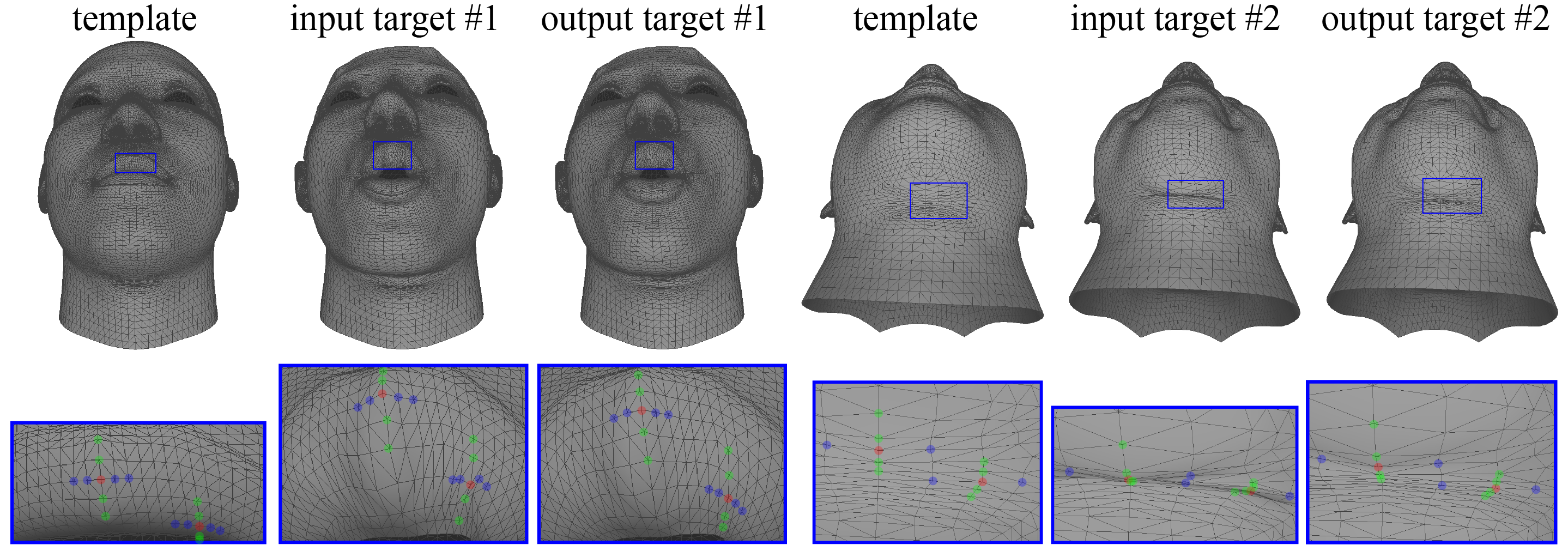}
	\end{center}
	\caption{Qualitative results for two examplar faces in FaceScape~\cite{yang2020facescape} dataset (filename: \textit{(122)13\_lip\_funneler.obj} and \textit{(340)4\_anger.obj}). The input target meshes are the initial results in correspondence with the template mesh by the NICP~\cite{amberg2007optimal} method, and the output targets are the optimized results by the proposed method in this work. Some corresponded vertices with notable differences by the proposed method from the input meshes are marked with colored dots for better viewing.}
	\label{fig1}
\end{figure*}
\IEEEPARstart{D}{ense} registration of 3D face seeks canonical representation of different facial surfaces such that their detailed structures can be compared. It is also named as dense correspondence in many prior works since the dense vertex-to-vertex mappings are established after the registration is completed. Dense registration of 3D face is fundamental to many downstream tasks for 3D and 2D facial analysis due to the invariance of 3D shape to pose and illumination variations. For example, it is an essential step for 3D statistical face modeling~\cite{blanz1999morphable,booth2018large,li2017learning,egger20203d,ferrari2017dictionary,gerig2018morphable,ploumpis2020towards}. It has also activated many solutions to problems for 2D faces~\cite{zhu2017face,garrido2014automatic,liu2016joint,bulat2017far,suwajanakorn2017synthesizing,zollhofer2018state,corneanu2016survey}.

Dense registration of 3D face belongs to the common non-rigid registration problem and has its own characters:
\begin{itemize}
	\item It is an ill-posed optimization problem since the solution is not uniquely defined. In a departure from the rigid case for solving an optimal rigid motion, the non-rigid registration problem belongs to a much larger class that has no explicit formulation.
	\item It is a domain-specific problem targeted at 3D facial surfaces. The intrinsic anatomical structure of face should be considered as vital clues for vertex-to-vertex correspondences to guide the non-rigid registration.
\end{itemize}

In the common case, the 3D shape (commonly surface in real-life applications) non-rigid registration problem~\cite{allen2003space,brown2007global,tam2012registration,ma2014robust,maiseli2017recent} can be revisited as an optimization problem for classical elastic shape similarity modeling~\cite{terzopoulos1987elastically,sorkine2007rigid}:

\emph{Given a template surface $\mathcal S$ and a target surface $\mathcal T$, the problem is to solve optimal parametrization that minimizes the following deformation energy measuring the difference between two surfaces as}

\begin{equation}\label{e1}
E(\mathcal S,\mathcal T) = \int_\Omega  {({k_s}\left\| {{\rm I}' - {\rm I}} \right\|_F^2 + {k_b}\left\| {{\rm I}{\rm I}' - {\rm I}{\rm I}} \right\|_F^2)} dudv.
\end{equation}
where is ${\left\|  \cdot  \right\|_F}$ is the Frobenius norm, ${\rm I}$ and ${\rm II}$ are the first and second fundamental forms of the surface $\mathcal S$, respectively, and ${\rm I}'$ and ${\rm II}'$ are the corresponding forms of $\mathcal T$ as the deformed version of $\mathcal S$. The first fundamental form measures the difference between $\mathcal S$ and $\mathcal T$ in a 2D embedding space on parameterized surface $\Omega$, while the second fundamental form measures the local difference of curvature. The two terms in Eq.~\ref{e1} are weighted by $k_s$ and $k_b$ for tangential and normal distortions, respectively. This formulation favors the non-rigid deformation to be locally rigid. However, local rigidity is very abstract and Eq.~\ref{e1} is difficult to be optimized directly. 

In a particular case for 3D face, the state-of-the-art works~\cite{amberg2007optimal,li2008global,myronenko2010point,pan2013establishing,ferrari2021sparse} commonly achieve vertex-to-vertex correspondence by a non-rigid deformation from a template face $\mathcal S$ to a target face $\mathcal T$. In a pairwise manner, the template face is usually a well-customized mesh as the initialization for registration. Then, the optimization for registration is generally modeled as an expectation–maximization (EM) problem as follows.
\begin{itemize}
	\item \textbf{E-step:} searching for a plausible preliminary vertex-to-vertex correspondence between $\mathcal S$ and $\mathcal T$.
	\item \textbf{M-step:} solving an optimal non-rigid deformation from $\mathcal S$ to $\mathcal T$ under some certain constraints.
\end{itemize}

The above process is in fact a generalization of the well-known iterative closest point (ICP) algorithm~\cite{besl1992method,chen1992object,zhang1994iterative} for rigid registration to the non-rigid case by alternating the above two steps.  In the E-step, correspondence of vertices is established with some certain rules, such as landmark correspondence as hard feature constraint and closest vertex correspondence as soft constraint. In the M-step, the non-rigid deformation is usually modeled as some locally smooth deformations, such as locally rigid transformation~\cite{fujiwara2011locally,fan2019boosting} and locally affine transformation~\cite{feldmar1996rigid,amberg2007optimal}. For example, the NICP~\cite{amberg2007optimal} algorithm includes a landmark term together with a distance and a stiffness term in the full objective function for optimization. The state-of-the-art (SOTA) works commonly include feature matchings for normal and curvature, \textit{e.g.} the iterative closest normal point (ICNP) methods~\cite{mohammadzade2012iterative}, therefore achieve reasonable results for minimizing the second term in Eq.~\ref{e1}. In this work, we show that the result can be further optimized for the first term of Eq.~\ref{e1}, which leads to better local rigidity, and furthermore a unique solution to the registration problem. This is neglected in the prior works yet a quite important issue for 3D face correspondence. In fact, better optimization for the first
term in Eq.~\ref{e1} results in superior representation of 3D facial surface. And it also leads to coherent local registration and elegant mesh grid routing referring to a template mesh as shown in Fig.~\ref{fig1}.

\begin{figure}[htbp]
	\begin{center}
		\includegraphics[width=1\linewidth]{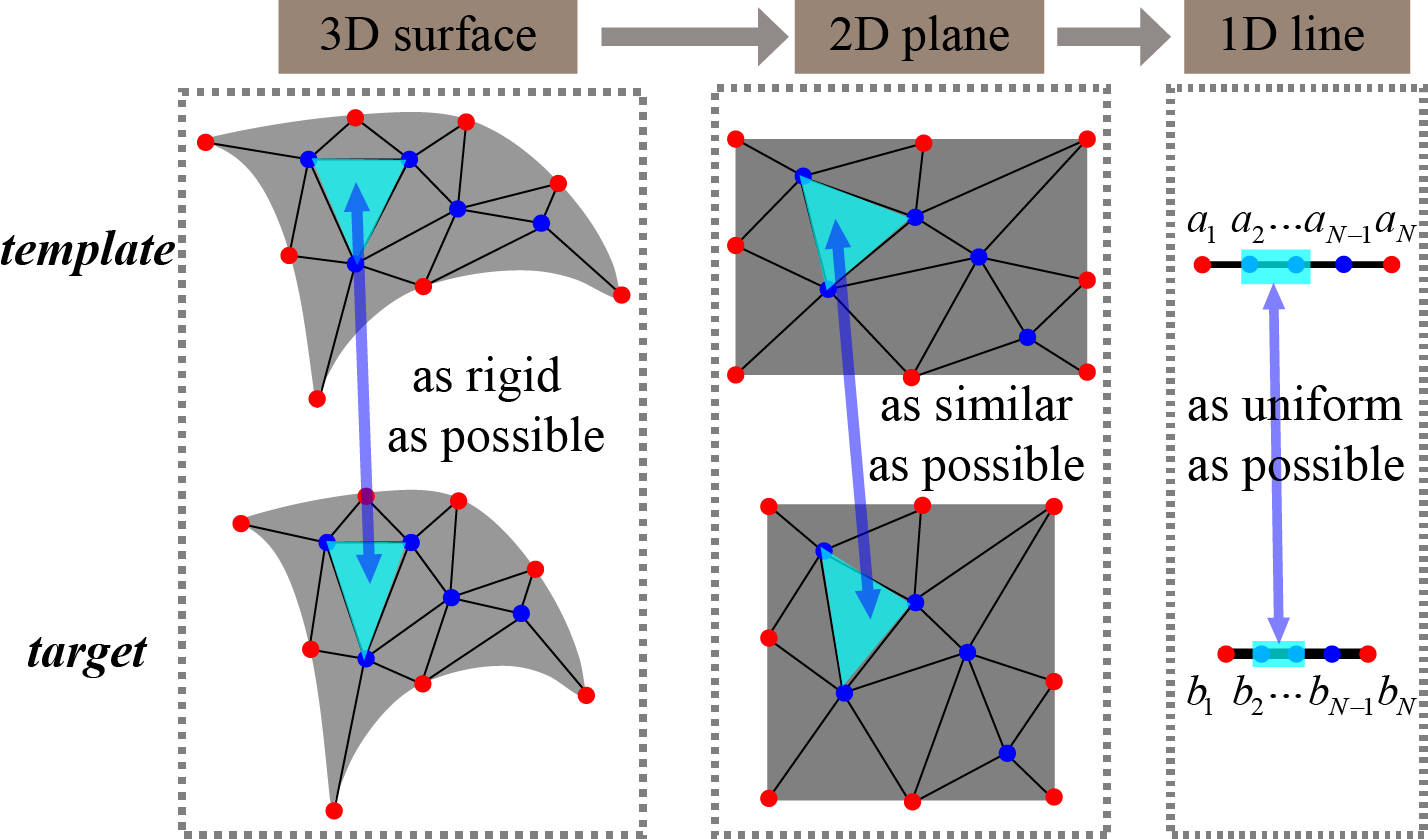}
	\end{center}
	\caption{Evolution from an original registration problem on a 3D surface to a 2D plane, and finally to a 1D line by the ``reasoning in lower dimension'' methodology.}
	\label{fig2}
\end{figure}

Optimizing the first term of Eq.~\ref{e1} for dense correspondence can be considered as dividing a surface according to a certain template, as in Fig.~\ref{fig2}. If we shrink the dimension of the original problem from 3D surface to 2D plane, then it becomes a problem to solve for optimal locations of vertices to divide the plane into similar triangles. If we further evolve the problem to the 1D case, then the definition is clear and the result is unique (as in Fig.~\ref{fig2}): 

\emph{Given a template line with end points $\{a_1,a_N\}$ and dividing points $\{a_2,...,a_{N-1}\}$, one can find optimal segmentations $\{b_2,...,b_{N-1}\}$ on a target line with end points $\{b_1,b_N\}$ by satisfying}
\begin{equation}\label{e2}
\frac{{{a_1}{a_2}}}{{{b_1}{b_2}}} = \frac{{{a_2}{a_3}}}{{{b_2}{b_3}}} =  \cdot  \cdot  \cdot  = \frac{{{a_{N - 1}}{a_N}}}{{{b_{N - 1}}{b_N}}}.
\end{equation}  

While the segmentation of a line in a 1D Euclidean space is simple, the generalization to a 3D surface on a Riemann manifold requires extensive studies. In this paper, we propose a novel dividing and diffusing method for 3D face dense registration. Starting from a target mesh in correspondence with a template mesh, the proposed method alternates between a dividing step and a diffusion step to achieve fine-grained vertex-to-vertex correspondence between them. We further propose a multi-resolution (MR) version of the dividing and diffusing method to facilitate the correspondence process. We show that the proposed method is correlated to the minimization of the distance between the template and the target face embedded in a logarithmic metric space for local scaling. In summary, the main contributions of this work are as follows:
\begin{itemize}
	\item We propose a novel dividing and diffusing method for fine-grained optimization of 3D face dense correspondence, which leads to superior representation of 3D facial shape.
	\item We propose a multi-resolution version of the dividing and diffusing method for fast convergence.
	\item We elucidate the physical meaning of the proposed method as smooth rearrangement of a local scaling metric for 3D facial shape, which leads to coherent local registrations and elegant mesh grid routines.
\end{itemize}

\section{Related Work}\label{sec:related_work}
3D face dense registration is an extremely non-linear optimization problem that requires a well initialization and plausible constraints for optimization. If we consider the problem as pairwise correspondence between two faces, then the optimization problem can be initialized by a template with a few labeled landmarks as feature correspondence to the target. Most of the current works in this field can be classified into deformation based and deformation free ones according to their optimization strategies.

\subsection{Deformation Based Method}
These methods commonly deform a template face to a target face under some constraints on reasonable initializations, feature correspondences, and locally smooth deformations, which are discussed respectively as follows. 

\textbf{Initialization.} A well-customized template mesh is generally used as the initialization for a subsequent deformation process. Then, the template face is usually aligned to the target face with a globally rigid or affine transformation driven by a few labelled landmarks or some other significant features in correspondence. For example, Mohammadzade \textit{et al.}~\cite{mohammadzade2012iterative} use automatically detected nose tip to guide the global alignment of the template face to the target face. Amberg \textit{et al.}~\cite{amberg2007optimal} use an affine transform to align the template face to the target face with the correspondence of a few annotated landmarks. Some other works~\cite{fan2018dense,li2008global,fujiwara2011locally} adopt rigid transformations to align the template to the target face. In particular, Fan \textit{et al.}~\cite{fan2019boosting} construct a group of locally rigid transformations to match the landmark locations exactly. The above methods generally employ rigid or approximately rigid registrations as the preliminary step for non-rigid deformation. 

\textbf{Feature correspondence.} Deformation guided by feature correspondence can maximally preserve the basic structure of a human face, while being able to converge to a reasonable solution for global optimization as well. Current works~\cite{amberg2007optimal,gilani2017deep,fan2018dense} commonly use some pre-annotated landmarks as feature correspondence across different faces. Generally, each pair of landmarks in correspondence dominates over other vertices in a local region around it. The landmark correspondences control the holistic shape and help to deal with expression variations in the guided deformation. There are also some other works that use landmark-free approaches in a departure from the landmark-based feature correspondences. For example, Gilani \textit{et al.}~\cite{zulqarnain2015shape} use level set geodesic curve to extract seed points automatically for global and local deformations. Fan \textit{et al.}~\cite{fan2018dense} use some high-entropy points, which can be considered as dense feature points in significantly high-curvature areas of face to guide the global deformation. Pan \textit{et al.}~\cite{pan2013establishing} use some denser points to model large deformations on the mouth region. The denser points in these methods can be regarded as robust representations of facial features, which are more tolerant to individual landmark matching errors. However, the capacity for guiding large deformation is discounted without accurate and definite landmarks, therefore being suboptimal for faces with large expressions to that with neutral expression. 

\textbf{Smooth deformation.} Locally smooth deformations also preserve the basic structures of a 3D facial surface, while enable the deformed template to adhere to the target face gradually. Myronenko and Song~\cite{myronenko2010point} regularize the offset of each vertex by total variation constraint for local smoothness, as a general non-rigid registration method. Patel and Smith~\cite{patel20093d} use a thin-plate spline warp towards the target face for smooth deformation. Zhang \textit{et al.}~\cite{zhang2016functional} incorporate functional maps into the deformation process, and ensure smooth local deformation by the low-frequency basis of the eigenfunctions of the Laplace-Beltrami operators. The NICP method~\cite{amberg2007optimal} and its variant~\cite{cheng2015active} incorporate locally affine transformations as smooth deformations. Fan~\textit{et al.}~\cite{fan2019boosting} construct a group of locally rigid transformations to guarantee smooth shape deformations. The deformations in these methods generally alternate with refined correspondence for each vertex on a face as restrictive optimization to acquire the results for non-rigid registration. In some other methods~\cite{vlasic2005tr2005,bolkart2015groupwise,ferrari2021sparse}, a prior model originated from a number of face prototypes in correspondence are incorporated to fit the deformed target faces. However, this is a chicken-and-egg problem and the correspondence problem of face prototypes remains to be solved. 

The deformation based methods are able to reach a reasonable solution with an alternative EM approach for minimizing Eq.~\ref{e1}. However, since most of these methods employ different rules for both surface deformation and vertex correspondence, they generally lead to distinct solutions and neglect the optimality of the results. Therefore, the 3D face dense correspondence problem has no standard solution in the existing works. In this work, we propose a dividing and diffusing algorithm to optimize the results for 3D face dense registration, in particular for the tangential parameterization in the first term of Eq.~\ref{e1}, and achieve stable results which are largely independent of preliminary correspondences.  

\subsection{Deformation Free Method}
Deformation free methods establish dense vertex-to-vertex correspondence without an explicit deformation process. These methods commonly involve some point matching strategies with the guidance of a few significant features, such as facial landmarks. Other geometric features such as curvature and normal are usually employed as signatures for local shape matching. We list some representative methods in the literature as follows.

In the seminal work of 3D Morphable Model (3DMM)~\cite{blanz1999morphable}, Blanz and Vetter propose to encode and map both 3D shape and texture features to a 2D cylindrical coordinate, and use a regularized optic-flow based algorithm for dense correspondence of 3D faces.

Sun and Abidi~\cite{sun2001surface} project geodesic contours around some feature points onto their tangential planes, which are further used as signatures for shape matching. This method is further improved and applied to 3D face dense correspondence by Salazar \textit{et al.}~\cite{salazar2014fully} on the BU-3DFE dataset~\cite{yin20063d} for modeling shape variation from large expressions.

Gu~\textit{et al.}~\cite{gu2007ricci,zeng2010ricci} propose a Ricci flow based method for conformal mapping of a 3D facial surface to a 2D canonical coordinate. Then, some uniform dividing strategies are conducted, with a few pre-annotated landmarks as fixed feature points.

Gilani~\textit{et al.}~\cite{gilani2017dense} first detect sparse correspondences on the outer boundary of a 3D face, and then triangulate existing correspondences and expand them iteratively by matching points of distinctive surface curvature along the triangle edges. Finally, the triangle mesh is refined by evolving level set geodesic curves.

The deformation free methods can be seen as employing some indirect dividing strategies on the target facial surface. Since the dividing of a 3D surface directly is difficult, some of these methods reduce the dimension of the original problem from 3D to 2D~\cite{blanz1999morphable,gu2007ricci,zeng2010ricci,grewe2016fully}. However, the limitation is that the intrinsic geometric features are only represented approximately, since it is unable to isometrically embed a non-flat 3D surface (\emph{e.g.} 3D face) into 2D plane perfectly. This is particularly difficult for complex 3D surface, such as 3D face with large expressions and 3D hand with flexible joints. In this work, we propose a novel iterative dividing and diffusing method on the original target facial surface directly referring to the mesh architecture of a template facial surface, in a departure from the existing indirect dividing methods. 

\section{Segmentation of a Line}\label{seg_line}
A scientific methodology to solve a problem in high-dimensional space is to first consider the problem in a low-dimensional case. Following this idea we cast the registration problem for 3D facial surface as 
a line segmentation problem shown in Fig.~\ref{fig2}. The problem asks for a tuple of optimal sub-lines $\{a_1 a_2,...,a_{N-1}a_{N}\}$ on a target line referring to a certain template line proportionately. One can simply use a ruler to measure the length of the two lines and then determine the dividing points according to Eq.~\ref{e2}. Unfortunately, such a global ruler measuring the ``length'' of a 3D surface does not exist, while the difference between two surfaces can be compared locally. Keeping in mind that we do not have a global ``ruler'', we employ an iterative approach to solve this segmentation problem instead. We formulate the problem as a dividing and diffusing process for solving Eq.~\ref{e2} as follows.

Given a template line $\mathcal A$ with points $\{a_1,...,a_N\}$ and a target line $\mathcal B$ with initialized corresponding points $\{b_1,b_2^{(0)},...,b_{N-1}^{(0)},b_N\}$, one can compute the optimal dividing points $\{b_2^*,...,b_{N-1}^*\}$ on $\mathcal B$ by iteratively alternating the following two steps:
\begin{enumerate}
	\item \textbf{Dividing.} Compute each optimal point $b_{i+1}(i = 1,2,...,N-2)$ on $\mathcal B$ referring to each sub-line triplet $a_ia_{i+1}a_{i+2}$ on $\mathcal A$ by satisfying 
	\begin{equation}\label{e3}
	\frac{{{a_i}{a_{i+1}}}}{{{b_i}^{(j)}{b_{i+1}^{(j+1)}}}} = 
	\frac{{{a_{i+1}}{a_{i+2}}}}{{{b_{i+1}^{(j+1)}}{b_{i+2}^{(j)}}}},
	\end{equation}
	where $j$ denotes the iteration number.
	
	\item \textbf{Diffusing.} Denoting $o_{i+1}=b_{i+1}^{(j+1)}-b_{i+1}^{(j)}$ as the renewing offset for each point $b_{i+1}$ in the $j^{th}$ iteration, a local average strategy is applied as 
	\begin{equation}\label{e4}
	o_{i+1}^*=(o_{i}+2o_{i+1}+o_{i+2})/4,
	\end{equation}
	and each renewed point is computed by
	\begin{equation}\label{e5} 
	b_{i+1}^{(j+1)}=b_{i+1}^{(j)}+o_{i+1}^*.
	\end{equation}
\end{enumerate}

\begin{figure}[htbp]
	\begin{center}
		\includegraphics[width=0.8\linewidth]{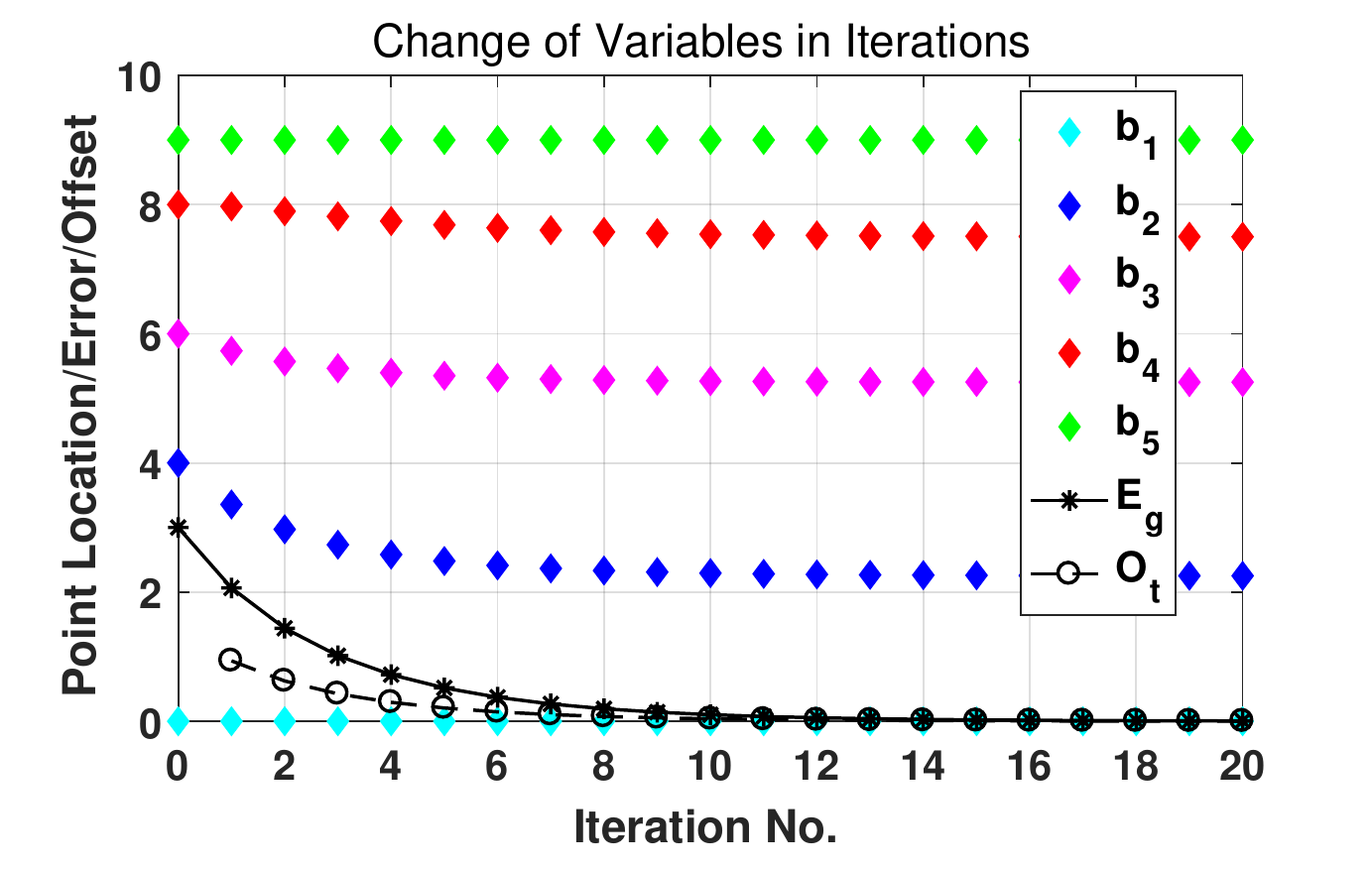}
	\end{center}
	\caption{Illustration of the iterative dividing and diffusing process in Table~\ref{table1}. The change of some variables is shown.}
	\label{fig3}
\end{figure}

The iterative process is able to reach a final solution when the renewed offset $\left|{o_{i+1}^*}\right|$ is smaller than a certain threshold. Table~\ref{table1} shows an example of the renewing process for a target line referring to a corresponding template line with $N=5$. The total error to the ground truth as
\begin{equation}\label{e6}
E_g=\sum\limits_{i = 2}^{N - 1} {\left| {{b_i} - b_i^*} \right|}, 
\end{equation} 
the total renewed offset as
\begin{equation}\label{e7}
O_t=\sum\limits_{i = 2}^{N - 1} {\left| {o_{i}^*} \right|}, 
\end{equation} 
and each renewing point $\{b_1,...,b_5\}$ are shown in Fig.~\ref{fig3}.

	\begin{table}
	\begin{center}
		\begin{tabular}{|c|ccccc|}
			\hline
			\multicolumn{6}{|c|}{\textbf{Iterative Line Segmentation w/o MR Scheme}}\\
			\hline
			\multirow{2}{*}{Iteration Number}&\multicolumn{5}{|c|}{Point Location}\\
			&$b_1$&$b_2$&$b_3$&$b_4$&$b_5$\\
			\hline
			0&$0$&$4$&$6$&$8$&$9$\\	
			1&$0$&$3.36$&$5.74$&$7.97$&$9$\\
			2&$0$&$2.97$&$5.57$&$7.90$&$9$\\
			5&$0$&$2.48$&$5.35$&$7.69$&$9$\\
			10&$0$&$2.29$&$5.27$&$7.54$&$9$\\
			18&$0$&$2.25$&$5.25$&$7.50$&$9$\\		
			\hline
			\hline
			\multicolumn{6}{|c|}{\textbf{Iterative Line Segmentation w/ MR Scheme}}\\
			\hline
			\multirow{2}{*}{Iteration Number}&\multicolumn{5}{|c|}{Point Location}\\
			&$b_1$&$b_2$&$b_3$&$b_4$&$b_5$\\
			\hline
			0&$0$&$4$&$6$&$8$&$9$\\	
			1&$0$&$4$&$5.25$&$8$&$9$\\
			2&$0$&$2.25$&$5.25$&$7.50$&$9$\\		
			\hline
			\hline
			Ground Truth&$0$&$2.25$&$5.25$&$7.50$&$9$\\	
		
			Reference Template&$0$&$3$&$7$&$10$&$12$\\	
			\hline
		\end{tabular}
	\end{center}
	\caption{
		An example of the iterative dividing and diffusing process for point correspondences. The points on the template line are $\{0,3,7,10,12\}$, and the points on the target line are initialized as $\{0,4,6,8,9\}$. It shows that the renewing points gradually approach to the ground-truth ones (directly computed via Eq.~\ref{e2}) as $\{0,2.25,5.25,7.50,9\}$ in $18$ iterations, while a multi-resolution scheme accelerates this process. 
	}
	\label{table1}
\end{table}

\begin{figure}[htbp]
	\begin{center}
		\includegraphics[width=0.8\linewidth]{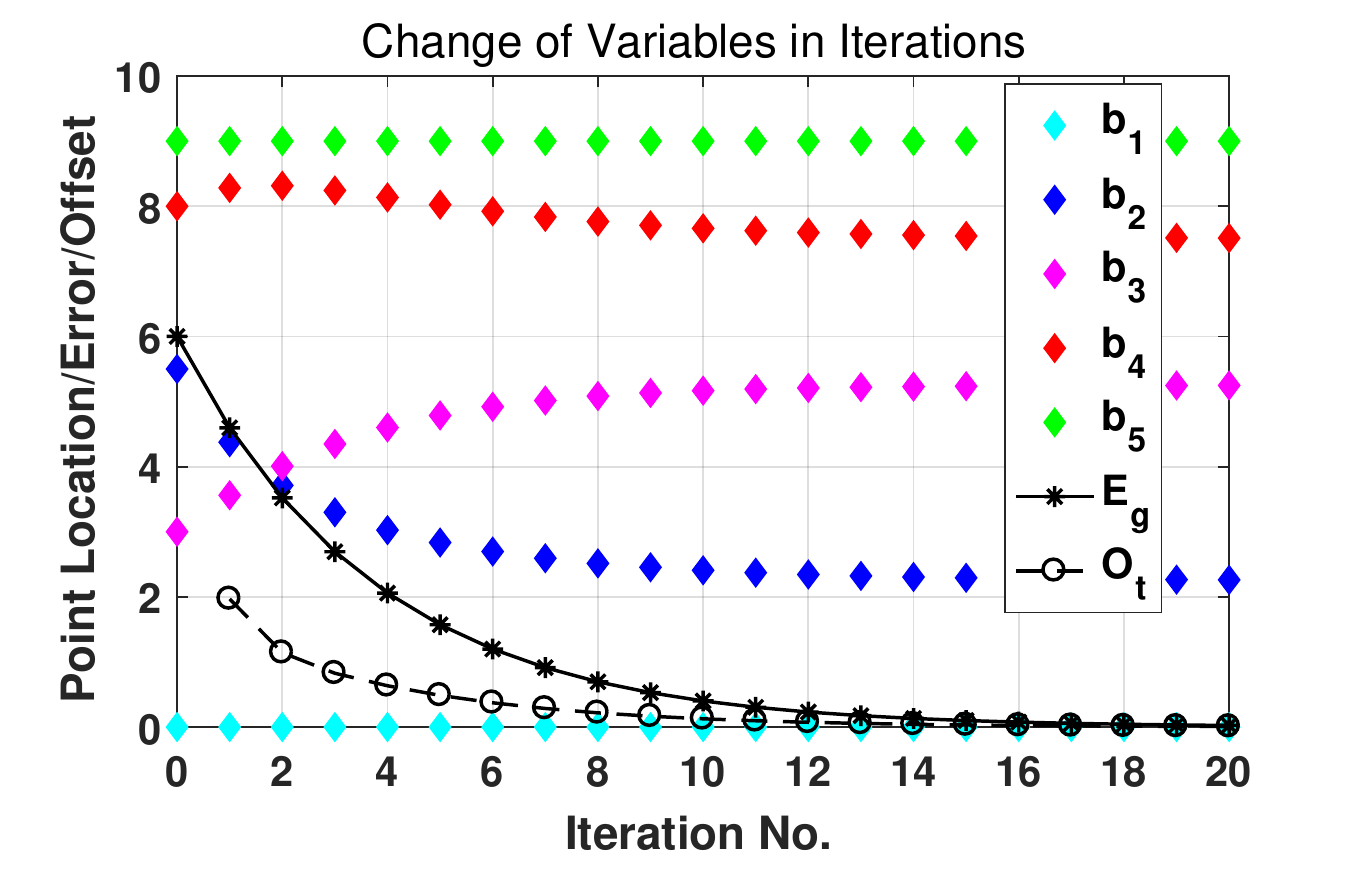}
	\end{center}
	\caption{An example for the iterative method dealing with self-intersected initialization. The points on the target line are initialized as $\{0,5.5,3,8,9\}$, with $b_2>b_3$ for self-intersection.}
	\label{fig4}
\end{figure}

This iterative process successfully achieves proportional division of a target line according to a certain template line. An additional advantage is its robustness to different initializations. The result still converges to the optimal solution, even if the initialized sublines intersect with each other, \textit{i.e.}, $b_i>b_{i+1},{\exists} \ i$. An example is shown in Fig.~\ref{fig4}, where the target line is initialized with $b_2>b_3$. This indicates that the generalized method on 3D surface can repair the self-intersection caused by flawed deformation, which is an attractive property for 3D face dense correspondence (also see Fig.~\ref{fig1}). However, the convergence speed is slow and can be improved further. If we use a multi-resolution scheme, \textit{i.e.}, first locate $b_3$ for a coarse resolution and then locate $\{b_2,b_4\}$ for a fine resolution, two iterations are sufficient for convergence as in Table~\ref{table1}. This motivates us to propose a multi-resolution (\textbf{MR}) version of the dividing and diffusing algorithm for 3D face dense registration, which will be elaborated afterwards.  

\section{Segmentation of 3D Facial Surface}\label{sec:DD_alg}
In this section, we propose an iterative dividing and diffusing method for the registration of 3D facial surfaces. Generalizing the algorithm in Section~\ref{seg_line} from 1D line to 3D surface requires appropriate definition of the ``dividing'' and ``diffusing'' process, respectively. We employ local registration of a vertex's $1$-ring neighbors as the local ``dividing'' process. We then construct a global optimization problem with constraints on local smoothing for the ``diffusing'' process. Contrary to the 1D case, the global optimization on a 3D surface involves additional fixed feature points (\textit{e.g.} landmarks) on the surface and a least-square formulation is put forward to deal with this issue.    

\subsection{Dividing}

The local registration process starts from a pair of initially registered template and target facial surfaces. We simply use the $1$-ring neighbors of a vertex as the basic local cell in the common triangle mesh representation of 3D surface. Other formats of data are also applicable by constructing a local cell with some nearest neighboring vertices. 

Let $\mathcal S=(\mathcal V^s, \mathcal E^s, \mathcal F^s)$ be a template mesh including $n$ vertices in $\mathcal V^s$, $m$ edges in $\mathcal E^s$, and $l$ triangles in $\mathcal F^s$; and let $\mathcal T=(\mathcal V^t, \mathcal E^t, \mathcal F^t)$ be the corresponding target mesh. For each vertex $v_i^s \in \mathcal V^s$, we denote its $1$-ring neighbors as ${\mathcal N}^1(v_i^s)$ and its corresponding vertices on the target as $v_i^t$ and ${\mathcal N}^1(v_i^t)$. Then, we suppose there exists a rigid transformation $\{R_i,T_i\}$ that aligns $v_i^s$ to $v_i^t$, as
\begin{equation}\label{e8}
{v_i^t} \leftarrow {R_i}{v_i^s} + {T_i}(v_i^s \in \mathcal V^s), 
\end{equation} 
where $\{R_i,T_i\}$ can be estimated by a least-square alignment problem of the surrounding $1$-ring neighbors, as 
\begin{equation}\label{e9}
\{ {R_i},{T_i}\}  = \mathop {\arg \min}\limits_{{R_i \in SO(3)},{T_i \in \mathbb R^3}} \sum\limits_{v_j^s \in {\mathcal N}^1(v_i^s)} {\left\| {{R_i}v_j^s + {T_i} - v_j^t} \right\|_2^2}.
\end{equation} 
Here $SO(3)$ denotes the space of all \emph{rank-$3$} orthonormal matrices with the determinants to be $1$ (\emph{i.e.} Givens Matrices). The problem in Eq.~\ref{e9} can be solved efficiently by a singular value decomposition (SVD) based method~\cite{sorkine2017least}. 

\begin{figure}[htbp]
	\begin{center}
		\includegraphics[width=1\linewidth]{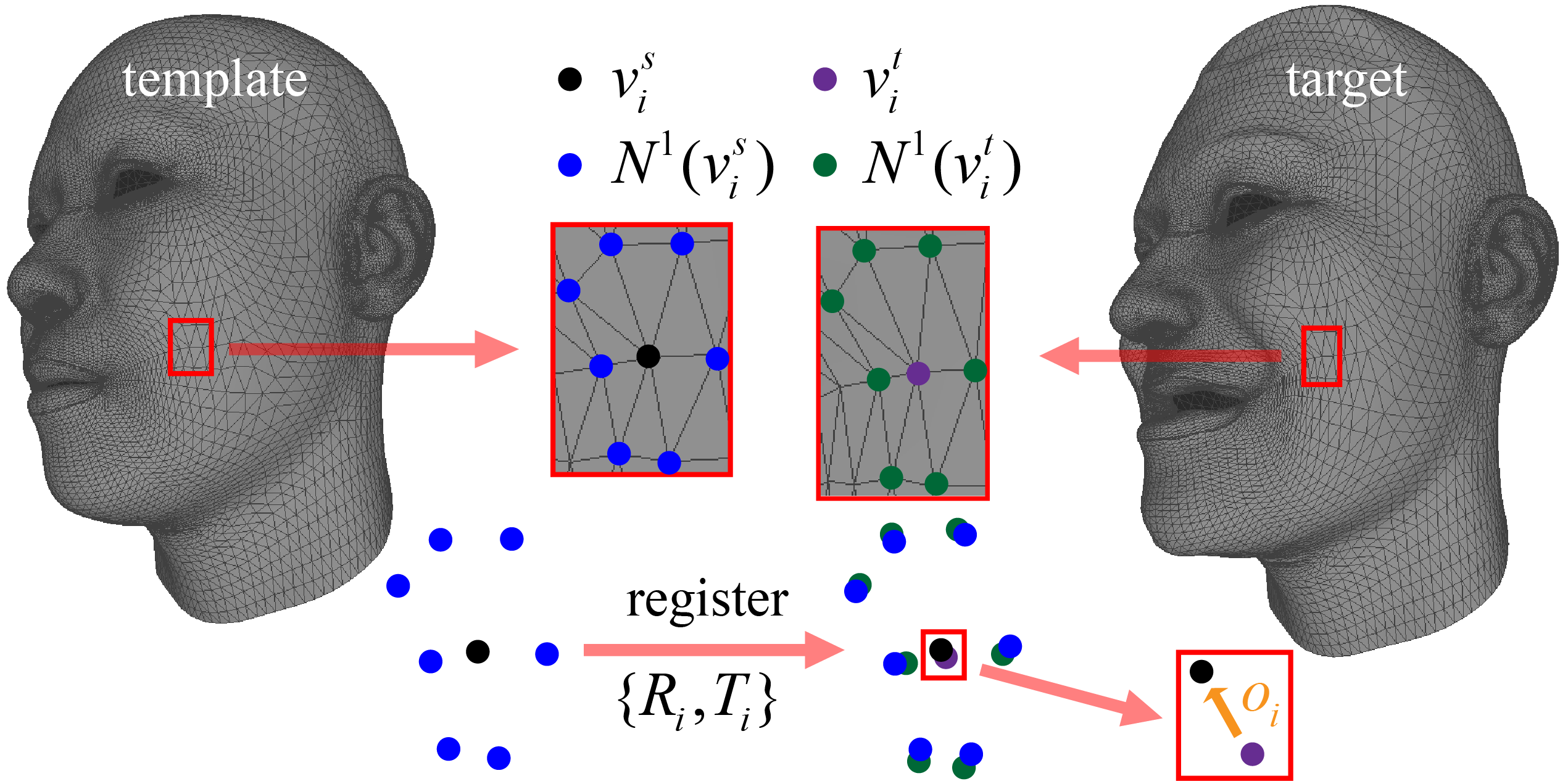}
	\end{center}
	\caption{Illustration of the local registration process for a vertex's $1$-ring neighbors.}
	\label{fig5}
\end{figure}

After the rigid transformation is obtained, we compute the preliminary offset for each vertex by 
\begin{equation}\label{e10}
o_i={R_i}{v_i^s} + {T_i}-{v_i^t}(v_i^s \in \mathcal V^s).
\end{equation}  
The local registration process is illustrated in Fig.~\ref{fig5}.

\subsection{Diffusing}
The preliminary offset for each vertex in Eq.~\ref{e10} should not be managed isolatedly, since the adjacent vertices share some common vertices in their $1$-ring neighbors. Therefore, we construct a diffusing (local smoothing) strategy to spread out the local effect for global optimization. We denote $p_i^t={R_i}{v_i^s} + {T_i}$ and formulate the problem as
\begin{equation}\label{e11}
\begin{aligned}
\begin{split}
\{ {o_i}|i \in \mathcal V\}  &= \mathop {\arg \min }\limits_{\{ {o_i}|i \in \mathcal V\} } \{ \sum\limits_{i \in \mathcal V} {\left\| {{p_i} - ({v_i} + o{}_i)} \right\|_2^2}\\
&+  \sum\limits_{i \in \mathcal V} (\lambda_i \cdot{\sum\limits_{j \in {\mathcal N}^1(v_i)} {\left\| {{o_i} - {o_j}} \right\|_2^2} )\} },
\end{split}
\end{aligned}
\end{equation} 
where $\lambda_i $ is the weights for diffusing of each offset and the superscript $t$ is omitted for brevity. 

Solving Eq.~\ref{e11} requires taking the partial derivative with respect to each offset $o_i$ and leads to a linear system
\begin{equation}\label{e12}
({[{{\bf{I}}_{ij}}]_{n \times n}}+{[{{\bf{A}}_{ij}}]_{n \times n}}) \cdot {[{{\bf{O}}_{ij}}]_{n \times 3}} = {[{{\bf{C}}_{ij}}]_{n \times 3}},
\end{equation}
where
\begin{equation}\label{e13}
{{\bf{A}}_{ij}} = \left\{ {\begin{array}{*{20}{c}}
	{\noindent \, {\lambda_i \cdot N_{v_i}}\quad \mbox{if}\quad i = j \,\,\qquad\qquad\qquad\,\,\,\,\,\,\,\,\,\,\,\,}  \\
	{\noindent \, -\lambda_i \,\,\,\,\,\,\,\,          \mbox{if}\,\,\,\,\,\,i \ne j \,\, \mbox{and} \, j \in {\mathcal N}^1(v_i)}  \\
	{\noindent 0 \quad\quad  \mbox{otherwise}\,\,\,\,\,\,\,\quad\quad\,\,\,\, \quad\quad}  \\
	\end{array}} \right.,
\end{equation}

\begin{equation}\label{e14}
{[{{\bf{O}}_{ij}}]_{n \times 3}} = {\left[ {{o}_1,...,{o}_n} \right]^T},
\end{equation}
and
\begin{equation}\label{e15}
{[{{\bf{C}}_{ij}}]_{N \times 3}}  = [p_1-v_1,...,p_n-v_n]^T.
\end{equation} 
Here $\bf{I}$ is the identity matrix, $N_{v_i}$ is the number of vertices in ${\mathcal N}^1(v_i)$, and the superscript $T$ denotes the operation for matrix transpose. If we let $\bf{B}=I+A$ be the coefficient matrix of Eq.~\ref{e12}, then $\bf{B}$ is a \emph{sparse} and \emph{strict diagonal-dominant matrix}. The solution for Eq.~\ref{e12} is discussed in two different cases:

\textbf{Case $\bf{1}$: Without Fixed Vertices.} In this case, all vertices are free to be renewed. Thus $\bf{B}$ is a \emph{full-rank square} matrix such that the linear system is well-posed and has a unique solution as
\begin{equation}\label{e16}
\bf{O}=\bf{B}^{-1}\bf{C}.
\end{equation}
This is similar to the mesh correction algorithm proposed in~\cite{fan2018dense}, which is applicable to neutral faces with small deformations. 

\textbf{Case $\bf{2}$: With Fixed Vertices.} In this case, some vertices are fixed as constraints for solving Eq.~\ref{e12}, in which $\bf{B}$ is a \emph{rank-deficient} matrix. Let $n_f$ be the number of fixed vertices ($\mathcal V_f \in \mathcal V$), we exclude the columns in $\bf{B}$ and the rows in $\bf{O}$ which are related to the fixed vertices. Then Eq.~\ref{e12} is equal to
\begin{equation}\label{e17}
{[{{\bf{B}}_{ij}}]_{n \times (n-n_f)}} \cdot {[{{\bf{O}}_{ij}}]_{(n-n_f) \times 3}} = {[{{\bf{C}}_{ij}}]_{(n-n_f) \times 3}},
\end{equation}
which is an \emph{over-determined} linear system. It can be further converted to a well-posed system as
\begin{equation}\label{e18}
{\bf{B}}^T {\bf{BO}}={\bf{B}}^T {\bf{C}}.
\end{equation}
Then the least-square solution to Eq.~\ref{e17} is
\begin{equation}\label{e19}
{\bf{O}}=({{\bf{B}}^T {\bf{B}}})^{-1}{\bf{B}}^T {\bf{C}}.
\end{equation}
By this way, we are able to process the mesh by fixing some already well-corresponded vertices while renewing other vertices. This is particularly useful for faces with large expressions, where feature points (\textit{e.g.} landmarks) are used to guide the overall correspondences. 

The vertex on the target mesh is added by each offset in $\bf{O}$ after solving Eq.~\ref{e11}, as
\begin{equation}\label{e20}
v_i^t=v_i^t+o_i(i\in \mathcal V/\mathcal V_f).
\end{equation} 
Finally, the closest vertex on the target surface to the vertex computed by Eq.~\ref{e20} is used to update the target mesh.

\subsection{The Overall Algorithm}
The overall algorithm alternates between the above dividing and diffusing process until the average of the renewed offset 
\begin{equation}\label{e21}
{\bar O_t} = \frac{1}{n-n_f}\sum\limits_{i \in \mathcal V/\mathcal V_f} {{{\left\| {{o_i}} \right\|}_2}}
\end{equation} 
is smaller than a certain threshold $\varepsilon$. In practice, the implementation of the proposed method involves the following details, which are not trivial.
\begin{figure}[htbp]
	\begin{center}
		\includegraphics[width=1\linewidth]{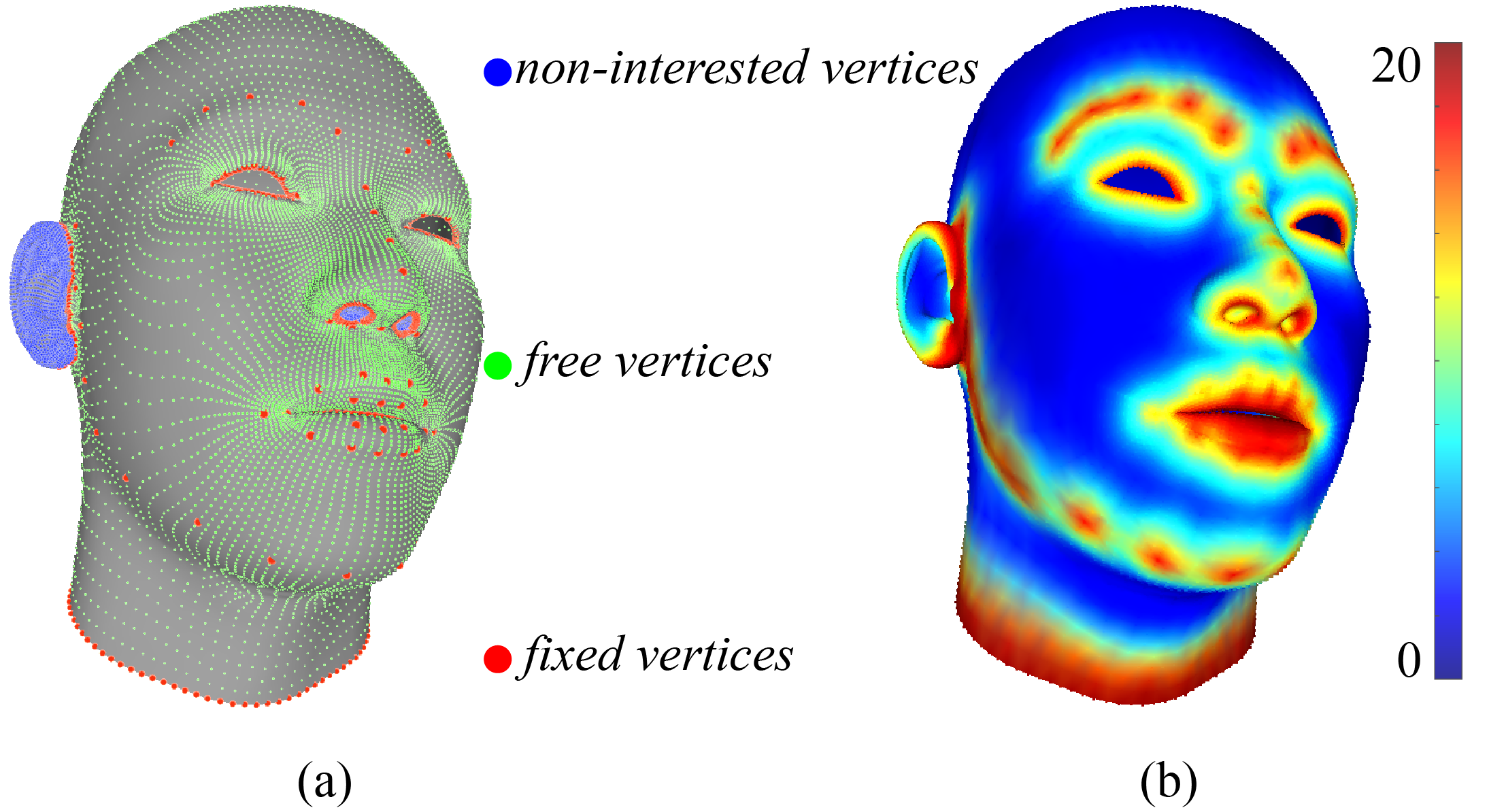}
	\end{center}
	\caption{(a) Different classes of vertices;(b) Weights for global diffusing.}
	\label{fig6}
\end{figure}

\begin{figure}[htbp]
	\begin{center}
		\includegraphics[width=0.6\linewidth]{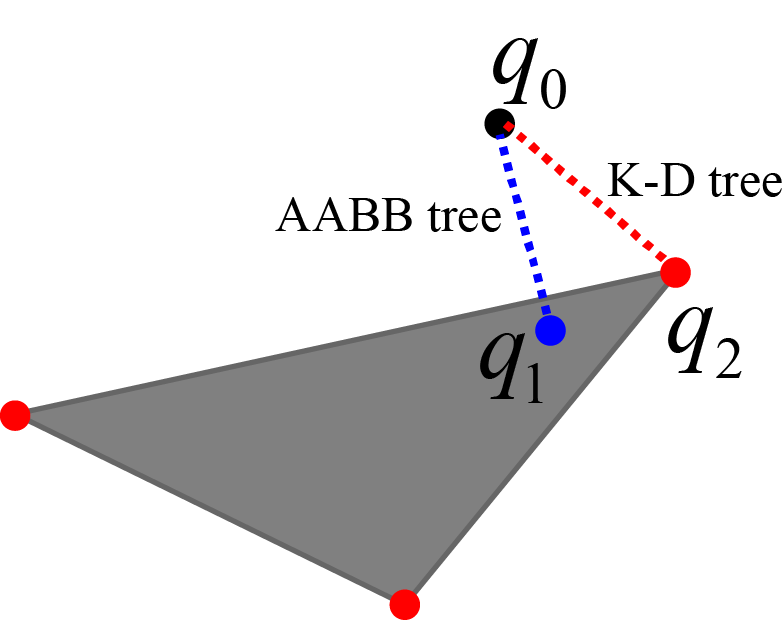}
	\end{center}
	\caption{The closest vertex searching on two different data structures. The closest vertex to a given vertex $q_0$ is $q_1$ and $q_2$ by the AABB tree and the K-D tree, respectively.}
	\label{fig7}
\end{figure}

\begin{itemize}
\item \textbf{Categorizing of Vertices.} The raw scanning data are commonly not able to capture some inner structures accurately on a face, such as the nostril and the ear regions. These structures are created from a template face in many 3D face dense registration method. We define the vertices in these regions as \emph{non-interested vertices}. The other vertices are defined as \emph{interested vertices} which are applied to the proposed method. In addition, the interested vertices can be classified into \emph{fixed vertices} and \emph{free vertices}. The fixed vertices include the landmarks, the edges of facial mesh, and the edges between interested and non-interested vertices. Fig.~\ref{fig6} (a) shows the different classes of vertices.

\item \textbf{Effects of Fixed Vertices.} The fixed vertices should serve as constraints during the dividing and diffusing process. We apply the dividing process for each fixed vertex and its $1$-ring neighbors. We then substitute each preliminary offset $o_i(i \in \mathcal V_f)$ by Eq.~\ref{e10} into the diffusing process. However, we do not renew the fixed vertices with Eq.~\ref{e20} in the iterations. Therefore, the fixed vertices are used to ``drag'' and ``pull'' the neighboring vertices to correct locations. Taking a row $i$ for a fixed vertex in Eq.~\ref{e17} for example, we obtain
\begin{equation}\label{e22}
\lambda_i \cdot \sum\limits_{j \in {\mathcal N}^1(v_i)} {{o_j}}  =  - ({p_i} - {v_i}),
\end{equation}  
where the fixed vertex $v_i$ transmits a ``reacting force'' to its neighboring vertices. To this end, we set the weights $\lambda_i$ for global diffusing in Eq.~\ref{e11} to decrease with the geodesic distance to the fixed vertices to enlarge the effect of fixed features, as in Fig.~\ref{fig6} (b). 

\item \textbf{Skills in Solving Eq.~\ref{e18}.} We do not use Eq.~\ref{e19} as matrix inversion to solve Eq.~\ref{e18} in practice. Since ${\bf B}^T{\bf B}$ is a symmetric positive definite matrix, we adopt three steps:

\textbf{I.} Re-ordering of variables and allocating memory;
 
\textbf{II.} Cholesky factorization of ${\bf B}^T{\bf B}$;
 
\textbf{III.} Substitution and iterative refinement.

\noindent Step \textbf{I} and \textbf{II} only require to be computed once in the whole algorithm. Therefore, solving step \textbf{III} in the sparse linear system is very fast by incorporating an advanced computational tool\footnote{https://pardiso-project.org/}.

\item \textbf{Nearest Neighboring Vertex Searching on Surfaces.} In recent works~\cite{pan2013establishing,yang2015go,gilani2017dense,fan2018dense} for nearest vertex searching on a 3D surface, a \textbf{K-D} tree architecture~\cite{bentley1975multidimensional} for point cloud is commonly used to accelerate the computational process. However, the closest vertex on a facial surface to a given vertex is not necessarily a vertex in a K-D tree. It generally locates insides a triangle unless the sampled point cloud is extremely dense. In this work, we use an \emph{axis-aligned bounding box} (\textbf{AABB}) tree~\cite{bergen1997efficient} considering both point cloud ($n$ vertices) and data structures ($l$ triangles) to improve the computational efficiency. This way also leads to accurate location of nearest neighboring vertex. Fig.~\ref{fig7} shows the difference between the results by the two different trees, which is commonly neglected in the existing works for 3D face dense registration. We use an advanced package\footnote{https://libigl.github.io/} for the implementation of AABB tree and it only requires to be built once for a certain target face in our method.
\end{itemize}

The overall method is concluded in Alg.~\ref{alg1}. This method can generally filter out the non-uniform mesh grid routine and achieve coherent local registrations for a target mesh, as shown in Fig.~\ref{fig1}. It also leads to better quantitative results for a group of data, which will be elaborated in the experiments.
\begin{algorithm}[htbp]
	\caption{Iterative Dividing and Diffusing on Surface} 
	\label{alg1}
	\hspace*{0.02in}{\bf Input:}\\ 
	\hspace*{0.02in}A template mesh
	$\mathcal{S}=(\mathcal{V}^s, \mathcal{E}^s, \mathcal{F}^s)$\\
	\hspace*{0.02in}A target mesh $\mathcal{T}=(\mathcal{V}^t, \mathcal{E}^t, \mathcal{F}^t)$\\
	{\bf Output:} \\
	\hspace*{0.02in}A target mesh $\mathcal{T}_{out}=(\mathcal{V},\mathcal{E}^t, \mathcal{F}^t)$
	\begin{algorithmic}[1]
		\State Initilize $\mathcal{V}=\mathcal{V}^t$ 
		\State Do re-ordering of variables, memory allocation, and matrix factorization for Eq.~\ref{e12}
		\State Build an AABB tree for $\mathcal{T}$
		\While{${\bar O_t}>\varepsilon$} 
		\State Compute each offset $o_i (i \in \mathcal V)$ by the dividing step in Eq.~\ref{e9} and Eq.~\ref{e10};
		\State Compute each regularized offset $o_i (i \in \mathcal V)$ by the diffusing step in Eq.~\ref{e11};
		\State Add each regularized offset $o_i (i \in \mathcal V)$ to $v_i (i \in \mathcal V)$ by Eq.~\ref{e20};
		\State Traverse the AABB tree on $\mathcal{T}$ for closest vertex searching and renew $v_i (i \in \mathcal V)$ accordingly.
		\EndWhile
		\State \Return $\mathcal{T}_{out}=(\mathcal{V},\mathcal{E}^t, \mathcal{F}^t)$
	\end{algorithmic}
\end{algorithm}

\section{Multi-resolution Analysis}\label{sec:DD_alg_MRA}
Section~\ref{seg_line} shows that a multi-resolution (\textbf{MR}) version of the dividing and diffusing algorithm on a line accelerates the convergence speed greatly. In this section, we generalize the multi-resolution based method to the facial surface and expect this will also benefit the convergence speed. To this end, the template face for registration is resampled into different resolutions. We incorporate a hierarchical strategy for the representation of 3D faces.

Since the proposed method starts from a template face and a topological uniform target face by an existing method,  multi-resolution analysis on the template face is sufficient. The counterpart on the target face follows the correspondence of each vertex. We organize the vertices on a template face in a pyramid structure with \emph{farthest point sampling} (\textbf{FPS}) method~\cite{eldar1997farthest}, which ensures uniform spacing of included vertex in an iterative manner. Suppose that a template face $\mathcal S=(\mathcal V^s,\mathcal E^s,\mathcal F^s)$ include $N_1$ free vertices and $N_2$ fixed vertices (original feature points), the steps for the reconstruction of a multi-resolution model are as follows.
\begin{enumerate}
	\item Compute vertex-to-vertex geodesic distances on $\mathcal S$.
	\item Define the number of free vertices as a pyramid architecture $\{N_1,N_1/4,...,N_1/4^{k-1}\}$ with $k$ different levels decimated by a factor $4$. 
	\item Initialize $N_2$ fixed vertices as included vertices for FPS.
	\item Include  $\{N_1/4^{k-1}, N_1/4^{k-2},...,N_1\}$ vertices sequentially using FPS, such that the MR model includes $\{N_1/4^{k-1}+N_2, N_1/4^{k-2}+N_2,...,N_1+N_2\}$ vertices for different resolutions $\{MR(0),MR(1),...,MR(k-1)\}$, respectively.	
\end{enumerate}

We use a fast heat-flow based method~\cite{crane2013geodesics} for the computation of vertex-to-vertex geodesic distance on the template mesh, which is the metric for FPS in our method. In the first coarse resolution $MR(0)$, we use the original feature points as fixed vertices and included vertices as free vertices. In other resolutions, we use the vertices in the previous resolution as fixed vertices, and the lately included vertices in the current resolution as free vertices. The dividing and diffusing algorithm is applied to each resolution in a cascade manner. Finally, we denote the method in Section~\ref{sec:DD_alg} as the \emph{full-resolution} $MR(k)$ and apply it to reach the final results. The different number of fixed and free vertices in each resolution is summarized in Table~\ref{table2}.
\begin{table}
	\begin{center}
		\begin{tabular}{|c|c|c|}
			\hline
			\textbf{Resolution}&\textbf{Free Vertices}&\textbf{Fixed Vertices}\\
			\hline
			$MR(0)$&$N_1/4^{k-1}$&$N_2$\\
			$MR(1)$&$N_1/4^{k-2}-N_1/4^{k-1}$&$N_2+N_1/4^{k-1}$\\
			$...$&$...$&$...$\\
			$MR(k-1)$&$N_1-N_1/4$&$N_2+N_1/4$\\
			$MR(k)$&$N_1$&$N_2$\\
			\hline
		\end{tabular}
	\end{center}
	\caption{
		The number of fixed and free vertices in different resolutions of the MR model. 
	}
	\label{table2}
\end{table}

\begin{figure*}[htbp]
	\begin{center}
		\includegraphics[width=0.9\linewidth]{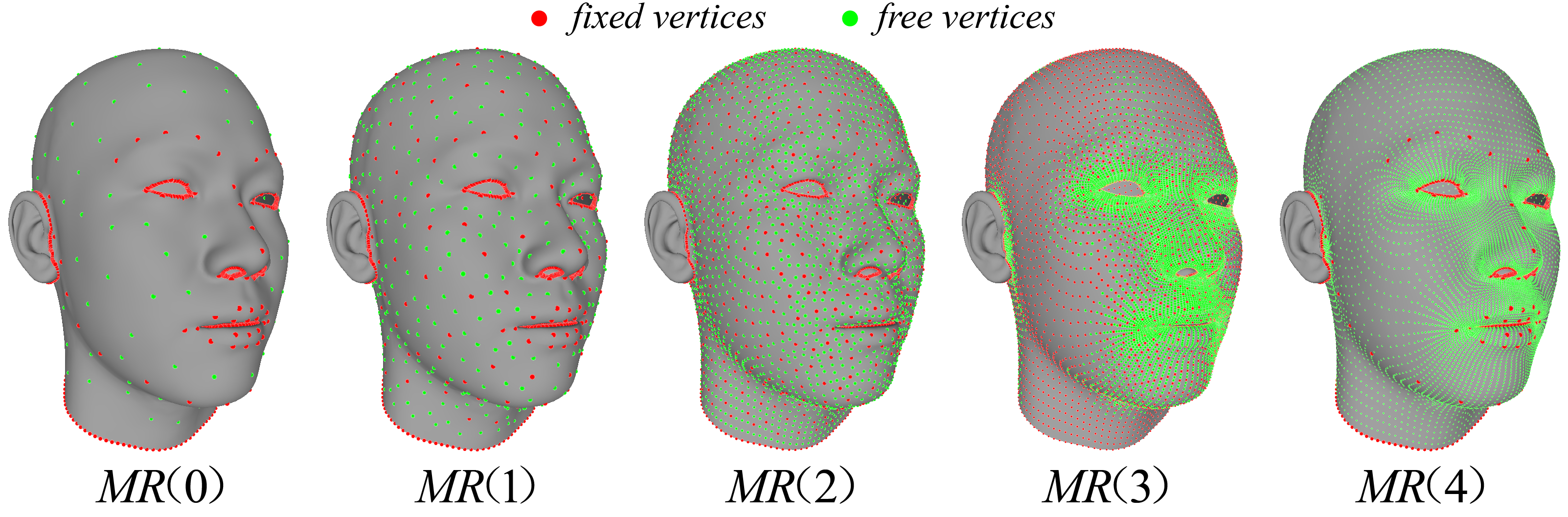}
	\end{center}
	\caption{An example of the customized multi-resolution model for a template face in the FaceScape dataset~\cite{yang2020facescape}.}
	\label{fig8}
\end{figure*}
An example for the MR model in the FaceScape dataset~\cite{yang2020facescape} is shown in Fig.~\ref{fig8}, where we set $k=4$. Unlike the full-resolution model that the neighboring relationship of all vertices are defined by the $1$-ring neighbors of the triangle mesh, this relationship for each vertex in other resolutions requires to be customized. In this work, we define the neighboring vertices by the following rules.
\begin{itemize}
	\item If a vertex is on the edges of the mesh, we adopt two vicinity vertices and another nearest vertex inside the mesh as its neighbors.
	\item If a vertex is inside the mesh, we adopt six nearest neighboring vertices as its neighbors.
	\item The neighboring relationship is symmetrized: \textit{i.e.} if vertex $a$ is incident to vertex $b$, then we make $b$ incident to $a$ as well.
\end{itemize}
In these rules, we use geodesic distance as the basic metric for nearest neighbor searching. The neighboring relationship is further used to replace the $1$-ring setting in Eq.~\ref{e11}. 


\section{Local Scaling Metric}\label{sec:Local_scaling_metric}
In this work, we consider the 3D face dense correspondence problem in a low-dimensional way, and propose an iterative dividing and diffusing method motivated by simple proportional segmentation of a line. As a result, the proposed method divides a target facial surface as rigidly as possible referring to a given template facial surface, while constrained by some fixed feature correspondences. In this way, the proposed method forces the difference between the target and template surface to vary in a locally smooth manner. We make some modifications upon Eq.~\ref{e2} and define a \emph{similarity score vector} between two faces according to their edge differences as
\begin{equation}\label{e23}
Q=[\frac{e_1^\mathcal T}{e_1^\mathcal S},\frac{e_2^\mathcal T}{e_2^\mathcal S},...,\frac{e_m^\mathcal T}{e_m^\mathcal S}],
\end{equation} 
where $e_i^\mathcal T$ and $e_i^\mathcal S$ are the corresponding edge lengths of the target and template face, respectively. The ratio $\frac{e_i^\mathcal T}{e_i^\mathcal S}(i \in \mathcal {E})$ is equal to $1$ if the corresponding edge lengths are the same. We define the \emph{local scale embeding vector} of a target face as 
\begin{equation}\label{e24}
S(\mathcal T)=[\log(\frac{e_1^\mathcal T}{e_1^0}),\log(\frac{e_2^\mathcal T}{e_2^0}),...,\log(\frac{e_m^\mathcal T}{e_m^0})]
\end{equation} 
by taking a common template face $\mathcal S=(\mathcal V^0,\mathcal E^0,\mathcal F^0)$ and using the logarithmic notation, where $e_i^0(i \in \mathcal E^0)$ is the length of each edge on the template face. By this way, the template face locates at the origin $[0,0,...,0]$ in a high-dimensional space. We define the induced \emph{local distance vector} between two faces $\mathcal T_1=(\mathcal V^1,\mathcal E^1,\mathcal F^1)$ and $\mathcal T_2=(\mathcal V^2,\mathcal E^2,\mathcal F^2)$ as
\begin{equation}\label{e25}
\begin{aligned}
d(\mathcal T_1,\mathcal T_2)&=|\mathcal S(\mathcal T_1)-\mathcal S(\mathcal T_2)|\\
&=[|\log(\frac{e_1^1}{e_1^0})-\log(\frac{e_1^2}{e_1^0})|,...,|\log(\frac{e_m^1}{e_m^0})-\log(\frac{e_m^2}{e_m^0})|]\\
&=[|\log(\frac{e_1^1}{e_1^2})|,...,|\log(\frac{e_m^1}{e_m^2})|],
\end{aligned}
\end{equation} 
where $e_i^k(i \in \mathcal E^k, k=1,2)$ are the lengths of the corresponding edges. We add up and average each element in Eq.~\ref{e25} and obtain a \emph{global distance metric} 
\begin{equation}\label{e26}
D: \mathcal T_1\times \mathcal T_2 \to \mathbb{R}
\end{equation} 
as
\begin{equation}\label{e27}
D(\mathcal T_1,\mathcal T_2) = \frac{1}{m}\sum\limits_{i = 1}^m {w_i\left| \log(\frac{e_i^1}{e_i^2}) \right|},
\end{equation} 
where 
\begin{equation}\label{e28}
w_i=\frac{\left|e_i^0\right|^2}{\sum\limits_{j=1}^m{\left|e_j^0\right|^2}}
\end{equation} 
is each normalized weight that compensates for different edge lengths on the template face. The metric in Eq.~\ref{e27} satisfies the following theorems.

\begin{figure*}[htbp]
	\begin{center}
		\includegraphics[width=1\linewidth]{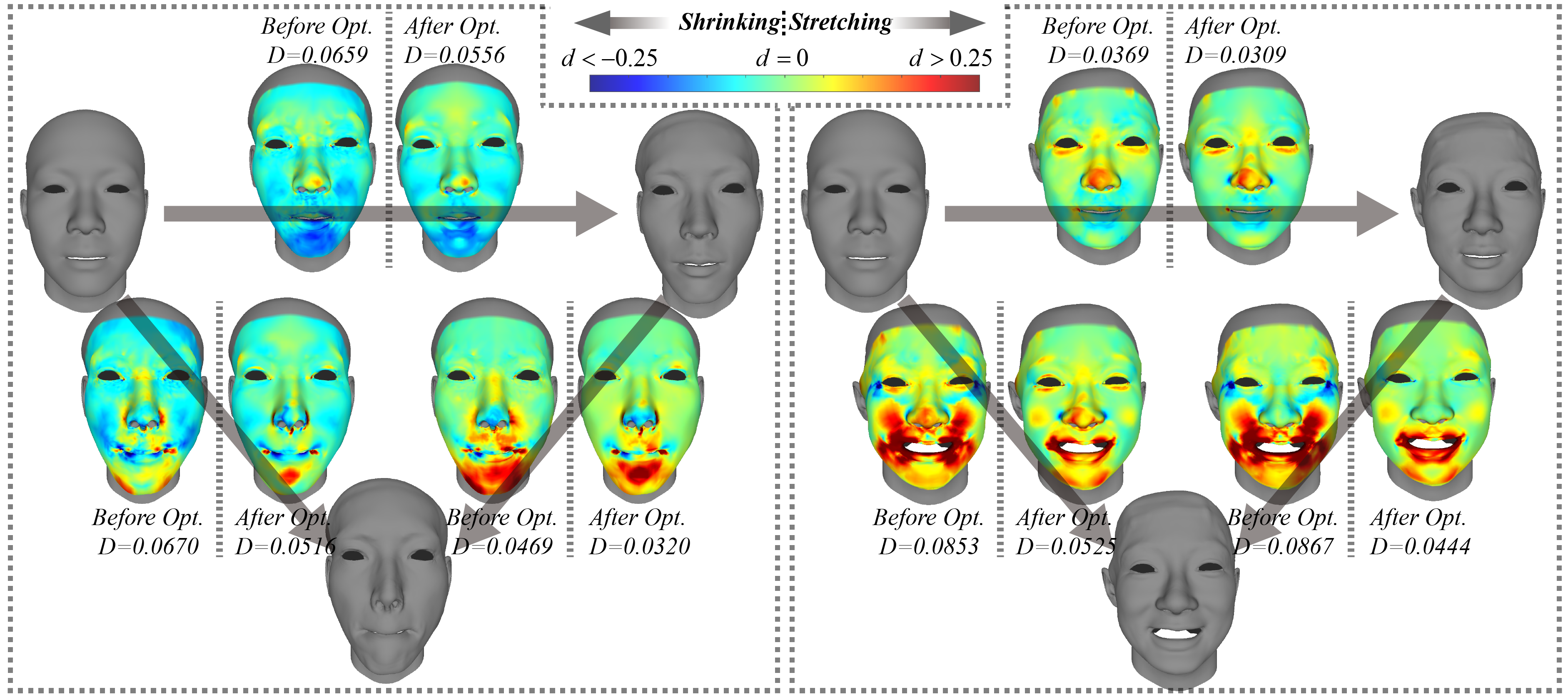}
	\end{center}
	\caption{Two examples in the FaceScape dataset~\cite{yang2020facescape}. The global distance metrics $D$ for each face before and after optimization by the proposed method are annotated, and the local distance vectors $d$ are illustrated as color maps on the facial surfaces. Please zoom in to see the more smooth and locally coherent details on the color maps after optimization.}
	\label{fig9}
\end{figure*}  
\noindent \textbf{Theorem I (Identity of Indiscernibles).} Given two arbitrary faces $\mathcal T_1$ and $\mathcal T_2$, we have
\begin{equation}\label{e29}
D(\mathcal T_1,\mathcal T_2) \ge 0,
\end{equation}
and the equality holds if and only if
\begin{equation}\label{e30}
\mathcal S(\mathcal T_1)=\mathcal S(\mathcal T_2).
\end{equation}

\noindent \textbf{Proof.} Inequality~\ref{e29} is correct since every element in Eq.~\ref{e27} is positive. We now focus on proving 
\begin{equation}\label{e31}
D(\mathcal T_1,\mathcal T_2)=0 \Leftrightarrow \mathcal S(\mathcal T_1)=\mathcal S(\mathcal T_2).
\end{equation}
The steps are as follows.
\begin{equation}\label{e32}
\begin{aligned}
&D(\mathcal T_1,\mathcal T_2)=0 \\
&\Leftrightarrow\frac{1}{m}\sum\limits_{i = 1}^m {w_i\left| \log(\frac{e_i^1}{e_i^2}) \right|}=0\\
&\Leftrightarrow[|\log(\frac{e_1^1}{e_1^2})|,...,|\log(\frac{e_l^1}{e_l^2})|]=[0,...,0]\\
&\Leftrightarrow[\frac{e_1^1}{e_1^2},...,\frac{e_l^1}{e_l^2}]=[1,...,1]\\
&\Leftrightarrow [e_1^1,...,e_l^1]=[e_1^2,...,e_l^2]\\
&\Leftrightarrow [\log(\frac{e_1^1}{e_1^0}),...,\log(\frac{e_m^1}{e_m^0})]=[\log(\frac{e_1^2}{e_1^0}),...,\log(\frac{e_m^2}{e_m^0})]\\
&\Leftrightarrow \mathcal S(\mathcal T_1)=\mathcal S(\mathcal T_2).
\end{aligned}
\end{equation}

\noindent \textbf{Theorem II (Symmetry).} Given the local scale embedding vectors $\mathcal S(\mathcal T_1)$ and $\mathcal S(\mathcal T_2)$ of two arbitrary faces, we have 
\begin{equation}\label{e33}
D(\mathcal T_1,\mathcal T_2)=D(\mathcal T_2,\mathcal T_1).
\end{equation}

\noindent \textbf{Proof.} Considering each of the elements in $D(\mathcal T_1,\mathcal T_2)$ and $D(\mathcal T_2,\mathcal T_1)$, we have
\begin{equation}\label{e34}
\left| \log(\frac{e_i^1}{e_i^2}) \right|
=\left|- \log(\frac{e_i^1}{e_i^2}) \right|
=\left| \log(\frac{e_i^2}{e_i^1}) \right|(i=1,...,m).
\end{equation} 
Thus Eq.~\ref{e33} holds.

\noindent \textbf{Theorem III (Triangle Inequality).} Given the local scale embedding vectors $\mathcal S(\mathcal T_1)$, $\mathcal S(\mathcal T_2)$, and $\mathcal S(\mathcal T_3)$ of three arbitary faces, we have 
\begin{equation}\label{e35}
D(\mathcal T_1,\mathcal T_3)\le D(\mathcal T_1,\mathcal T_2)+D(\mathcal T_2,\mathcal T_3).
\end{equation}
\noindent \textbf{Proof.} Considering each of the elements in $D(\mathcal T_1,\mathcal T_3)$, we have the following inequality
\begin{equation}\label{e36}
\begin{aligned}
&\left| \log(\frac{e_i^1}{e_i^3}) \right|\\
=&\left| \log(\frac{e_i^1 e_i^2}{e_i^2 e_i^3}) \right|\\
=&\left| \log(\frac{e_i^1}{e_i^2})+\log(\frac{e_i^2}{e_i^3}) \right|\\
\le& \left| \log(\frac{e_i^1}{e_i^2}) \right|+\left| \log(\frac{e_i^2}{e_i^3}) \right|(i=1,...,m).
\end{aligned}
\end{equation}
Then it follows
\begin{equation}\label{e37}
\begin{aligned}
D(\mathcal T_1,\mathcal T_3) &= \frac{1}{m}\sum\limits_{i = 1}^m {w_i \left| \log(\frac{e_i^1}{e_i^3}) \right|}\\
&\le \frac{1}{m}\sum\limits_{i = 1}^m {w_i(\left| \log(\frac{e_i^1}{e_i^2}) \right|+\left| \log(\frac{e_i^2}{e_i^3}) \right|)}\\
&=D(\mathcal T_1,\mathcal T_2)+D(\mathcal T_2,\mathcal T_3).
\end{aligned}
\end{equation}

Theorem I to III guarantee \emph{mathematical rigorism} for a given metric and indicate that the local scale embedding vector for each face can be treated as separable points in a high-dimensional Euclidean space. Physically, we represent the scale of a face locally referring to a template face in the tangential direction of the target surface. The geometric features of a face can be determined uniquely by the local scale in combination with the curvature (in the normal direction of the target surface). One merit of the local scale embedding vector in Eq.~\ref{e24} as a pack of scalars is its invariance to rigid transformations, thus represents the intrinsic features for a face. The proposed dividing and diffusing algorithm tends to minimize the global distance metric for two faces denoted by Eq.~\ref{e27}. We show two examples in Fig.~\ref{fig9} for both the local and global metrics in Eq.~\ref{e25} and Eq.~\ref{e27}. We can see that the global distance metric by Eq.~\ref{e27} is significantly smaller after applying the proposed method between two faces, while the vector in Eq.~\ref{e25} becomes more smooth and locally coherent on the target surface. In Fig.~\ref{fig9}, we can also see that the local distance vector measures the degree of stretching ($d>0$) or shrinking ($d<0$) of a face with respect to another face. This is particularly meaningful as an indicator for shape variances caused by different identities and expressions.

\section{Experiment}
Assessing the results for 3D face dense registration is not an easy task. On the one hand, some feature vertices are definite for anatomical significances and are visibly salient for local shape and texture features. On the other hand, most vertices for a dense face model are not definite with the ground-truth. Some existing works commonly use indirect ways for evaluation under different perspectives, such as representation ability of resulted face model~\cite{fan2019boosting,gilani2017dense,ferrari2021sparse}, landmark localization accuracy~\cite{segundo2010automatic,creusot2013machine,gilani2017deep,liu20193d,ferrari2021sparse}, and 3D face recognition performance~\cite{bronstein2005three,wang2010robust,drira20133d,fan2018dense}. In this work, the proposed dividing and diffusing method gives practicable definition to the problem by extending the fixed feature correspondence to the overall dense correspondence. Thus we treat 3D landmark detection (either manually or automatically annotated) as a pre-processing step and do not use it for evaluation. We do not use 3D face recognition performance for evaluation either, since it requires post-processing, such as shape clipping and feature extraction fed into different classifiers. In addition to the commonly used metric as groupwise representation ability of resulted face model, we include extensive experiments including computational efficiency, robustness to initialization, and direct metric embedding tailored for the proposed method for a full evaluation in different perspectives.

\subsection{Datasets} 
We carry out the experiments on two publicly available datasets including \textbf{BU-3DFE}~\cite{yin20063d} and \textbf{FaceScape}~\cite{yang2020facescape}. They are two typical datasets since both of them are rich for expressions and densely constructed by scans from multiple directions. BU-3DFE includes $2,500$ facial samples with $6$ different expressions in $4$ different levels by $100$ actors. The resolution of raw scanning data is around $10,000$ vertices per face. FaceScape is publicly available recently in $2020$ and includes $18,760$ high-resolution (around $2,000,000$ vertices) 3D faces with $20$ different expressions. It also provides the registered faces which share the same topology as triangle meshes with $26,317$ vertices and $52,261$ triangles. It is worth mentioning that the collection of large and high-quality 3D dataset is not an easy task. The existing works on 3D dataset~\cite{savran2008bosphorus,phillips2005overview,cao2013facewarehouse,booth2018large} play indispensable roles for the advancement of 3D face applications.  

\begin{figure}[htbp]
	\begin{center}
		\includegraphics[width=1\linewidth]{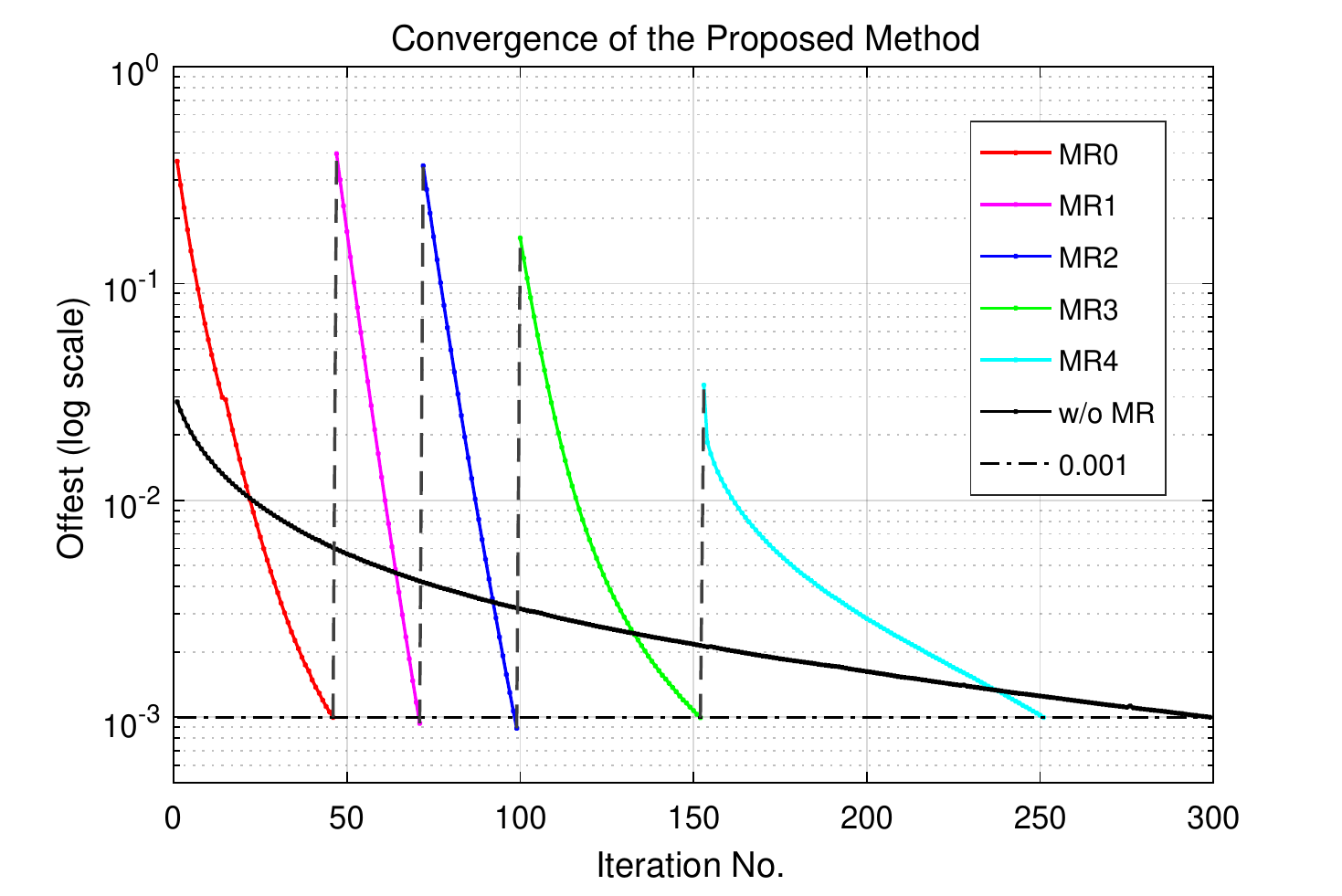}
	\end{center}
	\caption{An example for convergence speed of the proposed method. The full resolution version converges in $299$ iterations, while the MR version converges in $46$,$25$,$28$,$53$, and $99$ iterations for different resolutions in a cascade manner.}
	\label{fig10}
\end{figure}

\subsection{Computational Efficiency \& Convergence Analysis}
In the proposed method, the sparse matrix decomposition, the neighboring relationship of each vertex, and the multi-resolution organization of vertices require to be computed only once for a common template face. The building of AABB tree is computed once for each target face in the registration process. Thus we neglect the cost for these steps. The local registration process and the iterative refinement for solving Eq.~\ref{e17} contribute to most of the computational time in practice. The former involves small matrix multiplications ($3 \times {\mathcal N_{v_i}}$) and SVD decomposition ($3 \times 3$) for each vertex (totally $n$), while the latter involves iterative substitution of a sparse  rank-$(n-n_f)$ coefficient matrix. Thus the computational complexity for the full-resolution version is approximately
\begin{equation}\label{e38}
\begin{aligned}
C_{Full}&=o(n)+o(n-n_f)\\
&=o(n_f)+o(n-n_f).
\end{aligned}
\end{equation} 
In a MR version with the number of vertices growing exponentially in each resolution, the computational complexity is theoretically reduced to 
\begin{equation}\label{e39}
C_{MR}=o(n_f)+o(\log (n-n_f)).
\end{equation} 
Thus the computational time increases linearly and logarithmically with the number of vertices for the full resolution and the MR version of the proposed method, respectively. In our implementation\footnote{We implement the proposed method on the software and hardware environments of VC++2015 and i7-9700 (single thread), respectively.}, it requires $\bf{13.56s}$ for the full-resolution version on average for a model with $26,317$ vertices, while the MR model reduces the time to $\bf{5.85s}$. Fig.~\ref{fig10} shows an example by setting the threshold in Eq.~\ref{e21} to $\bf{0.001}$, which is normalized by the mean edge length of a template face. We include both the full-resolution version and the MR version for comparison. It shows that the proposed method is very efficient, while the MR version accelerates the convergence speed further. The faster convergence of the MR version is due to both less iterations and less vertices for operations in low-resolution steps. 

\begin{figure}[htbp]
	\begin{center}
		\includegraphics[width=0.85\linewidth]{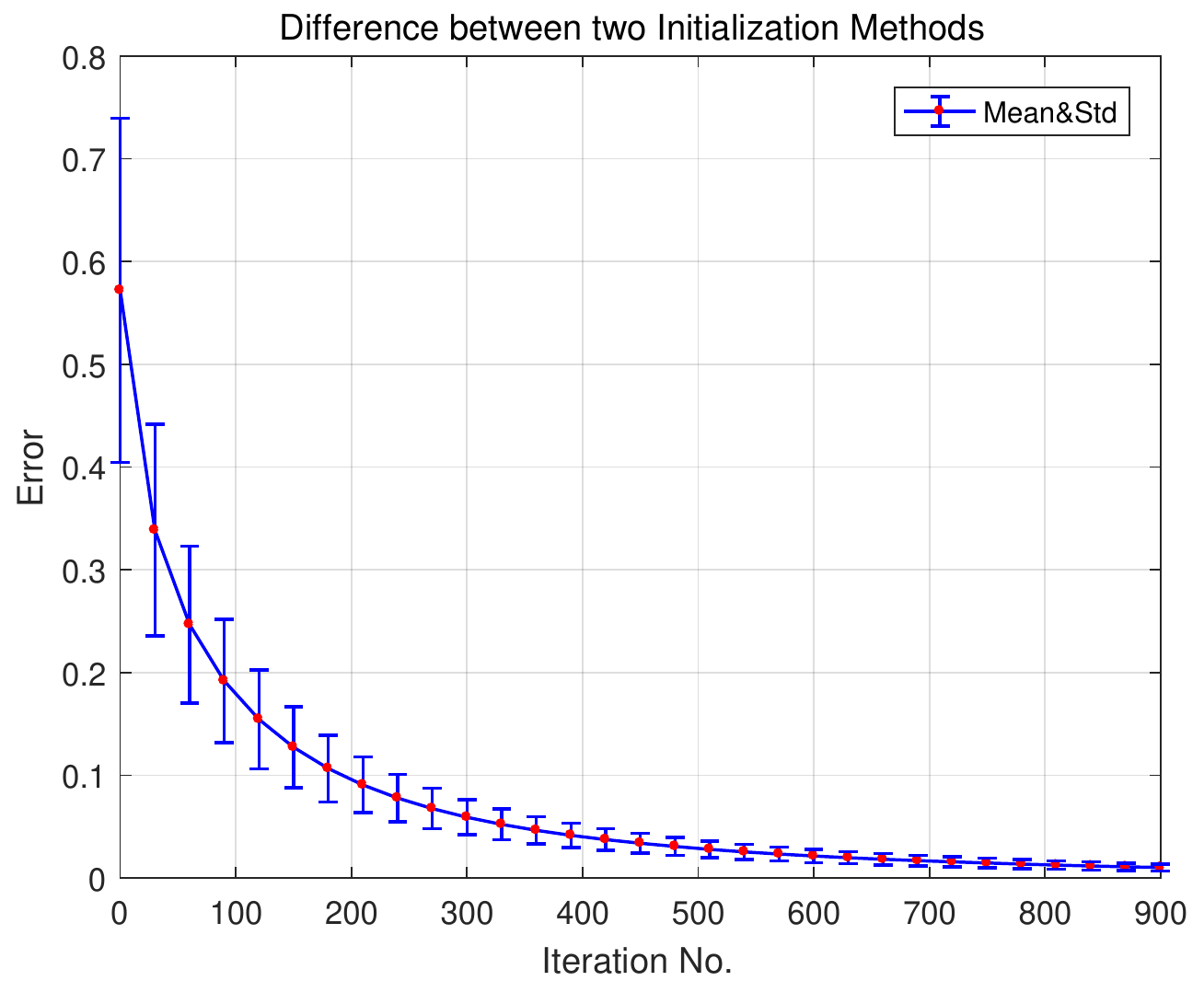}
	\end{center}
	\caption{The differences for per-vertex errors between the optimized results from two different initialization methods as functions to the iteration steps. The errors for both the mean and standard deviation are normalized to $[0,1]$ with a common factor in the errorbar map.}
	\label{fig11}
\end{figure}

\subsection{Definite with Stable Correspondence}
Section~\ref{seg_line} shows an example for iterative line segmentation which results in a unique solution defined by Eq.~\ref{e2}. We further guess that the proposed dividing and diffusing method is also able to converge to a unique solution which is independent to different initializations. We now test the influence of different initialization methods and noisy initializations, respectively. 

\begin{figure*}[htbp]
	\begin{center}
		\includegraphics[width=0.9\linewidth]{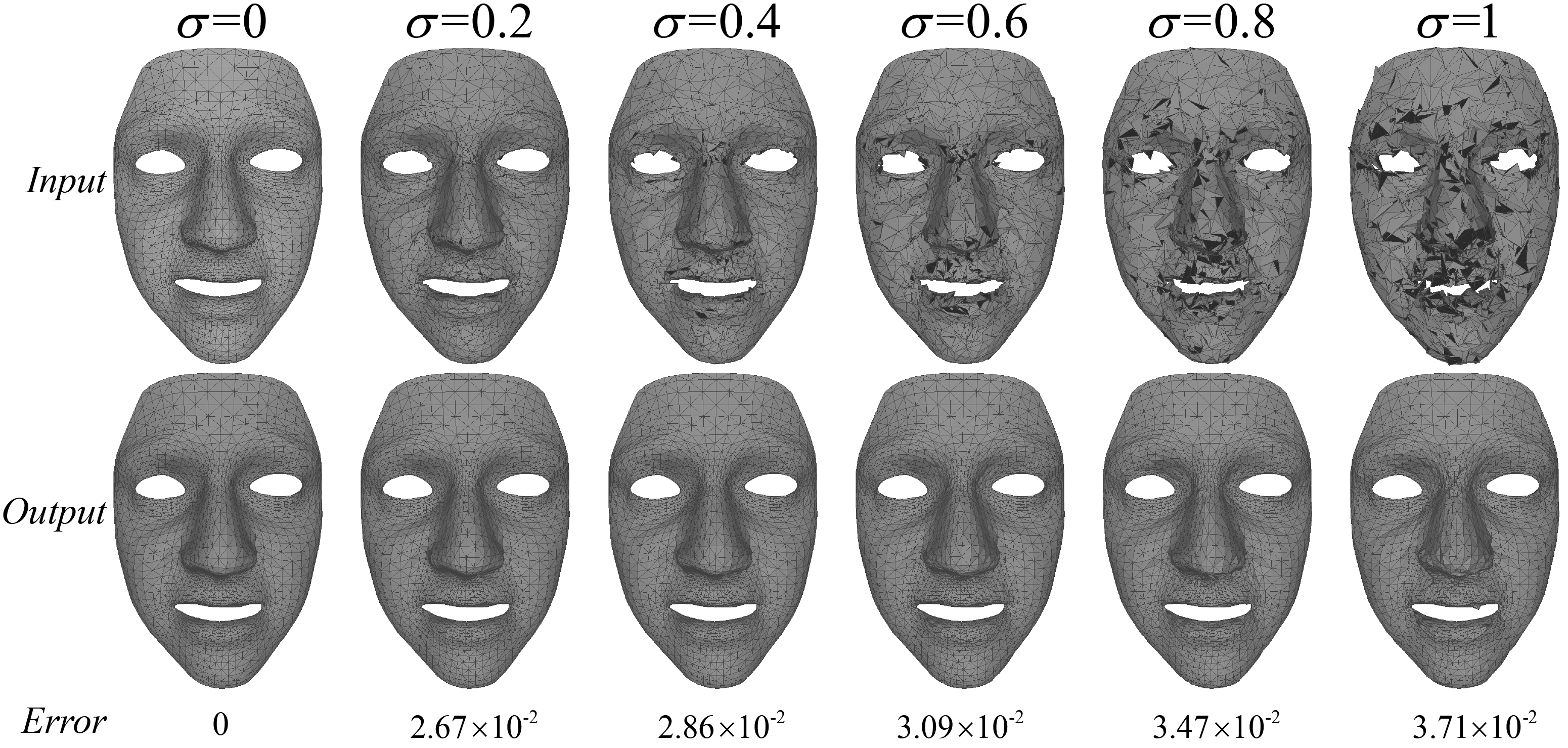}
	\end{center}
	\caption{Qualitative results for an exemplar face with different level of noise in the BU-3DFE dataset. Both the average per-vertex errors and the standard deviations of added noise are normalized by the mean edge length of a template face. Please zoom in to see the details for mesh grid routine.}
	\label{fig12}
\end{figure*}

\textbf{Initialization with different methods.} First, we select $100$ samples in the BU-3DFE dataset which cover different identities and expressions. The initial correspondences of these samples are established with two different and reproducible methods, the NICP method~\cite{amberg2007optimal} and a local shape deformation (LSD) method~\cite{fan2019boosting}. We apply the proposed method (full-resolution version) to these corresponding samples and compare the average per-vertex errors between the results from the  initializations with the two different methods. Fig.~\ref{fig11} shows the normalized results. We can see that both the mean and the standard deviation of the errors decrease with the number of iterations. This means that the results by the proposed method are independent from different initializations to some extent. The final residual errors should be caused by different mesh-edge vertices (as fixed vertices) and shape fitting errors in the normal direction by different methods.  

\textbf{Initialization with noise.} Then, we extract the corresponded faces in the BU-3DFE dataset by the local shape deformation method~\cite{fan2019boosting} and add different level of noise on free vertices for each face in the tangential directions. The distribution of noises on each vertex follows a uniform distribution with variance $\sigma^2$. Fig.~\ref{fig12} shows a qualitative example of optimized meshes with different levels of added noise. The average per-vertex errors between the results from a clean target and that with noise are also annotated. We can see that the proposed method is able to ``repair'' the noisy mesh grid routine and leads to locally coherent registrations. The vertex locations of corresponding faces and the mesh structures on the results across different level of noise are also stable. Therefore, the proposed method is very robust to noisy initializations.

The above experiments on different initializations show some evidences that the proposed method leads to stable correspondences with the fixed vertices (fixed features and mesh-edges). Therefore, the proposed method gives the definition of vertex correspondences on smooth regions of faces to some extent, which is a critical issue in the field of 3D face dense correspondence.

\subsection{Evaluation with the Proposed Metric}
Section~\ref{sec:Local_scaling_metric} introduces a local scaling metric that forces the registered target face to vary in a locally smooth manner referring to a template face. We now use both the global (Eq.~\ref{e27}) and local (Eq.~\ref{e25}) annotations for this metric to evaluate the correspondence results for the FaceScape and BU-3DFE dataset. We choose the topological uniform samples provided by the FaceScape dataset as initializations for the proposed method. The dense correspondences of these samples in topological uniform formats are achieved by a classical NICP~\cite{amberg2007optimal} method carefully with extra manual assistance. Therefore, we consider this as an enhanced NICP (\textbf{E-NICP}) baseline. The implementation of non-rigid registration methods commonly involves a lot of parameter settings and additional expertise in this field, against which the public baseline for FaceScape dataset avoids the induced subjective biases. In addition, we implement both the NICP method~\cite{amberg2007optimal} and the local shape deformation method~\cite{fan2019boosting} as baselines for the BU-3DFE dataset. The parameters for both of the two methods are carefully tuned to minimize implementation errors.

First,  we apply the proposed method (both the full-resolution and MR versions) to the $16,900$ samples\footnote{We exclude some problematic data in this dataset.} in the FaceScape dataset and $2,500$ samples in the BU-3DFE dataset. Table~\ref{table3} shows the results for the global metric in Eq.~\ref{e27} averaged over all samples in comparison with those by the baseline methods. We can see that the global metric is smaller after applying the proposed method, which denotes better coherent local deformations. In addition, the results by the full-resolution version and the MR version show no significant differences, since the full-resolution optimization is adhered to the final step of the MR version in this work.

\begin{figure*}[htbp]
	\begin{center}
		\includegraphics[width=0.9\linewidth]{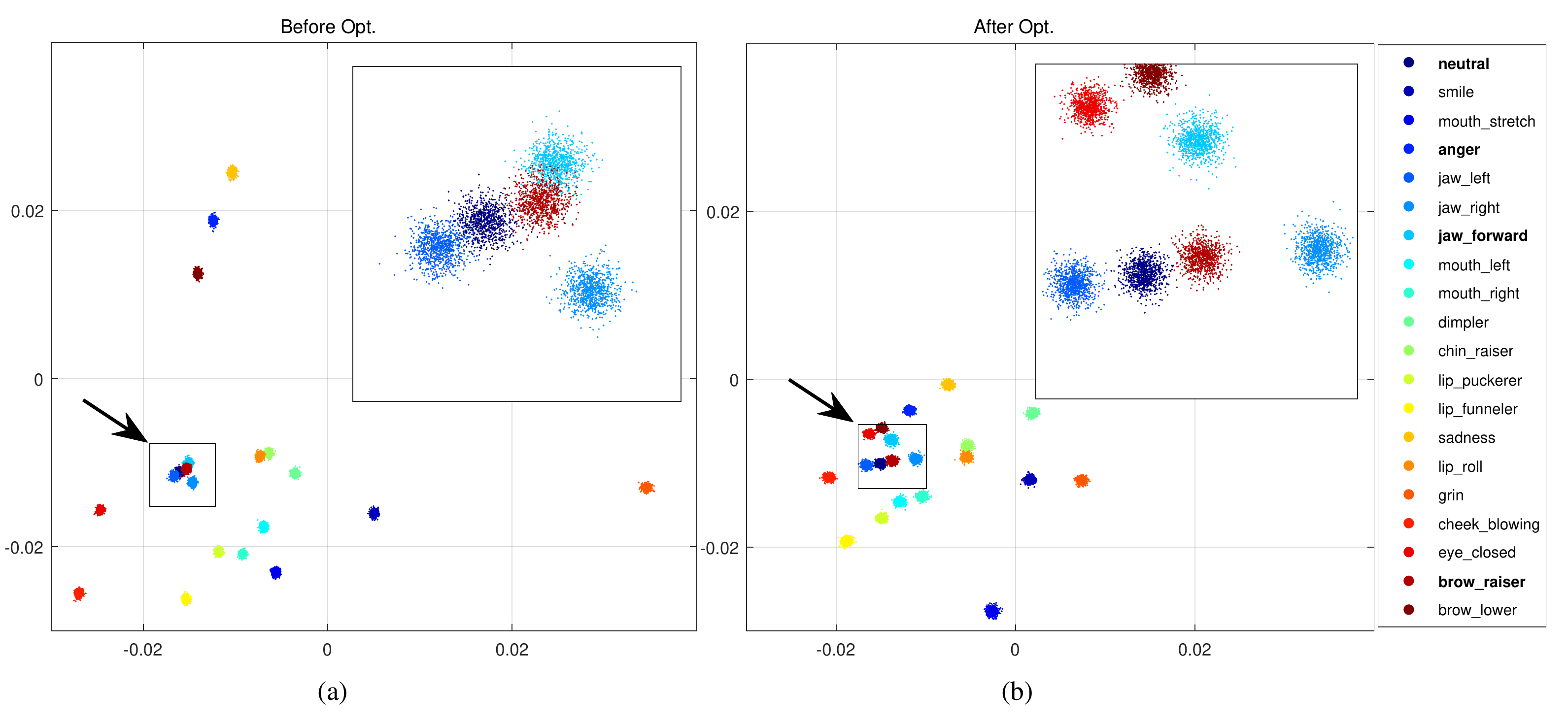}
	\end{center}
	\caption{Clustering results by LDA with respect to $20$ different expressions for $16,900$ samples on the FaceScape dataset. The $4$ mixed expressions are marked with arrows on the maps and bold typeface on the legend.}
	\label{fig13}
\end{figure*}

\begin{figure}[htbp]
	\begin{center}
		\includegraphics[width=1\linewidth]{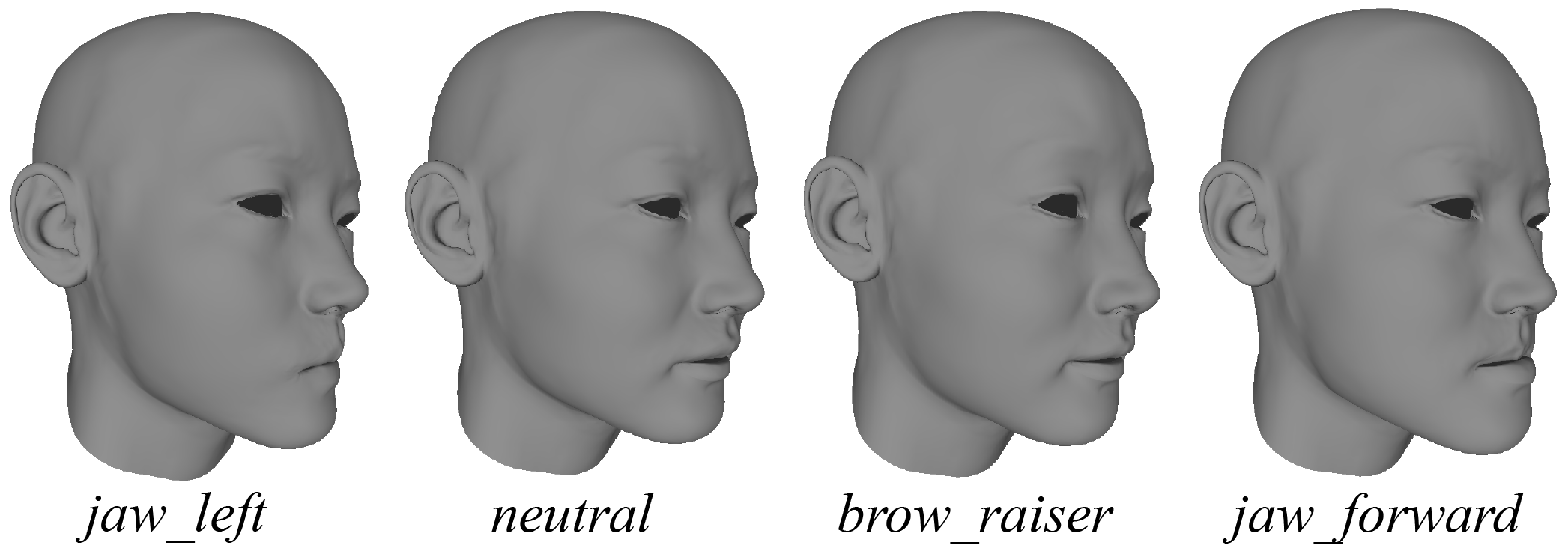}
	\end{center}
	\caption{An example for faces (\textit{ID: 395}) with $4$ different expressions.}
	\label{fig14}
\end{figure}

Then, we apply linear discriminative analysis (\textbf{LDA}) to the local scaling embedding vectors of $16,900$ samples with respect to $20$ different expression labels. We show the clustering results for the first $2$ dimensions before and after applying the proposed method in Fig.~\ref{fig13} (a) and Fig.~\ref{fig13} (b), respectively. It shows that the discriminate ability with respect to different expressions is largely improved after applying the proposed method. Specifically, the expressions with \textit{jaw\_left, neutral, brow\_raiser, and jaw\_forward} (see Fig.~\ref{fig14} for an example) are mixed sequentially in pairs with each other for original representations with the E-NICP method, whereas they are separated after optimization by the proposed method. This demonstrates that the proposed method leads to better discriminative representation for 3D facial shape, which is beneficial to many downstream applications. 
\begin{table}[htbp]
	\begin{center}
		\begin{tabular}{|c|c|cc|}
			\hline
			Dataset&FaceScape~\cite{yang2020facescape}&\multicolumn{2}{|c|}{BU-3DFE~\cite{yin20063d}}\\
			\hline
			Baseline Method&E-NICP~\cite{yang2020facescape}&NICP~\cite{amberg2007optimal}&LSD~\cite{fan2019boosting}\\
			\hline
			Before Opt.&$0.061$&$0.058$&$0.055$\\
			\hline
			After Opt.(Full)&$0.044$&$0.042$&$0.041$\\	
			\hline
			After Opt.(MR)&$0.045$&$0.042$&$0.042$\\	
			\hline
		\end{tabular}
	\end{center}
	\caption{Average quantitative results for the global metric on $16,900$ samples of the FaceScape dataset and $2,500$ samples of the BU-3DFE dataset before and after the proposed optimization method. The (enhanced) NICP and the local shape deformation (LSD) method are used as baselines for the initializations of the correspondence results.}
	\label{table3}
\end{table}


\begin{figure*}[htbp]
	\begin{center}
		\includegraphics[width=1\linewidth]{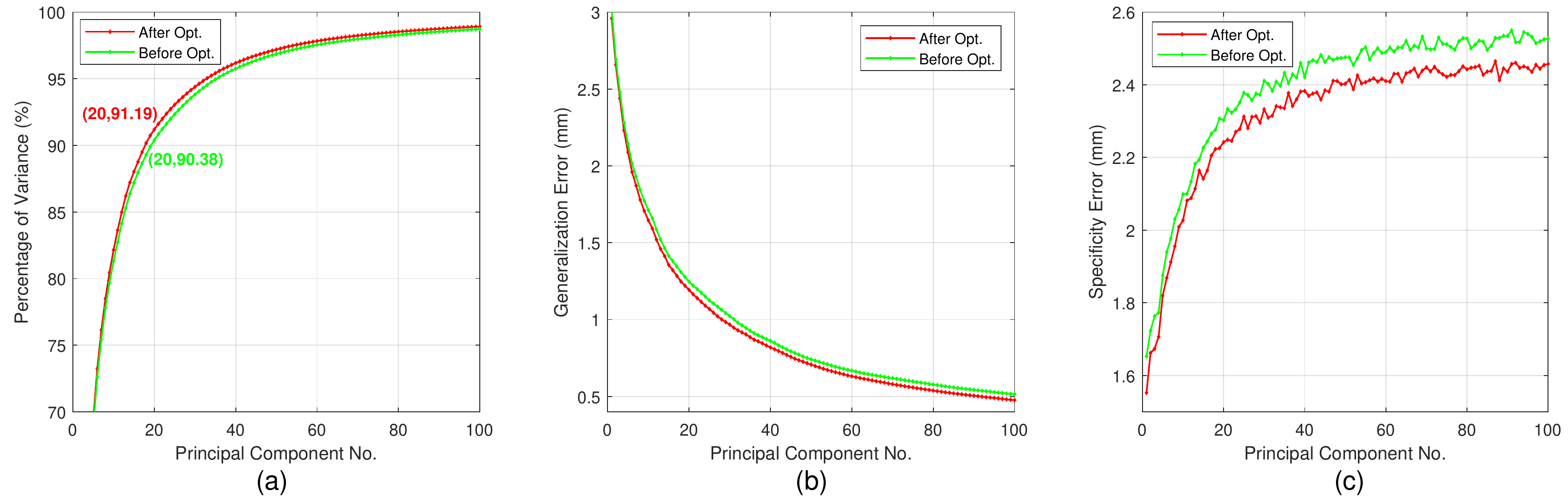}
	\end{center}
	\caption{(a) Compactness, (b) generalization, and (c) specificity for low-dimension representations before and after applying the proposed method.}
	\label{fig15}
\end{figure*}

\subsection{Compactness, Generalization, and Specificity for Low-dimensional Representations}
Fig.~\ref{fig1} shows that the proposed method leads to locally coherent results between the template mesh and the target mesh qualitatively. We hope the pairwise locally coherent results also benefit the global groupwise representations of 3D facial shapes. We follow a common practice~\cite{davies2002minimum} for the evaluation of statistical shape models using  \emph{compactness}, \emph{generalization}, and \emph{specificity}. We use the topological uniform data in the FaceScape dataset as standard baseline for the E-NICP method.

\textbf{Training\&Test Set Split.} The 3D faces in the FaceScape dataset include $938$ identities with $20$ typical expressions. We include the first $400$ identities ($7,995$ samples) as the training set and the rest $538$ identities ($8,905$ samples) as the test set. We apply principal component analysis (\textbf{PCA}) to the aligned training samples to achieve a low-dimensional representative model referring to the 3DMM~\cite{blanz1999morphable} pipeline. 

\textbf{Compactness} is measured by the percentage of variance retained by the PCA model. Compactness is a vital property of data for dimension reduction. It is evaluated by a \emph{minimum description length} principle~\cite{rissanen1978modeling} in the machine learning community as ``\emph{the less, the better}''. We compare the PCA models constructed from samples before and after applying the proposed methods, respectively. The results are shown in Fig.~\ref{fig15} (a). The first $20$ principal components explain 90.38\% and 91.19\% of the variances by the baseline model and that achieved by the proposed method, respectively. This demonstrates better compactness of low-dimensional models by the proposed method, which indicates accurate correspondence of 3D faces. 

\begin{figure}[htbp]
	\begin{center}
		\includegraphics[width=1\linewidth]{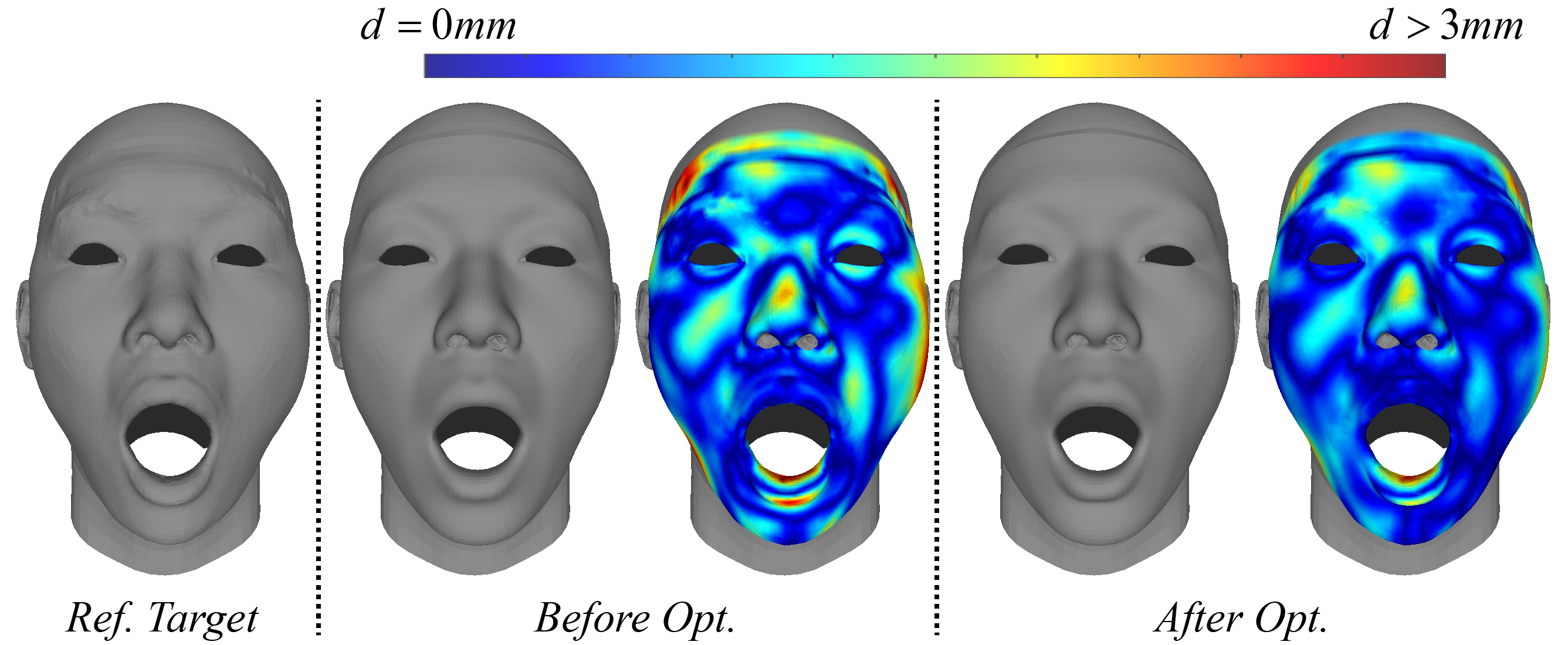}
	\end{center}
	\caption{An exemplar face (filename: \textit{(594)3\_mouth\_stretch.obj}) showing the difference between the generalization errors before and after applying the proposed method.}
	\label{fig16}
\end{figure}

\textbf{Generalization} measures the ability of a model to represent novel facial shapes that are not included in the training samples. We use the low-dimensional models, which are constructed with the training samples before and after applying the proposed method, respectively, to represent the test samples with different number of principal components. The mean vertex representation errors (as per-vertex $\mathcal L_2$ distances) averaged over all samples are shown in Fig.~\ref{fig15} (b). We observe that the generalization error is reduced notably after applying the proposed method to the corresponding training samples. Fig.~\ref{fig16} also shows a qualitative result for the representation of an exemplar facial sample with $20$ principal components, where there are visible differences between the models before and after applying the proposed method. These results demonstrate that the proposed method leads to not only better mesh grid routines (as in Fig.~\ref{fig1}) but also enhanced ability to represent shape variances.

\textbf{Specificity} measures the validity of generated faces by a representative model. We assume each principal component of the model follows Gaussian distribution with certain variances and randomly generate novel instances for a fixed number of principal components according to each Gaussian distribution model. The generated samples are then used to search their closest samples on the test set with minimum Euclidean distances. Fig.~\ref{fig15} (c) shows the average per-vertex distances with $1,000$ generated samples for different settings of principal component number. It shows that the results by the proposed method achieve smaller specificity error compared with the baseline method. This demonstrates the effectiveness of the generative model for synthesizing novel facial samples.

\subsection{Failure Cases}
\begin{figure}[htbp]
	\begin{center}
		\includegraphics[width=1\linewidth]{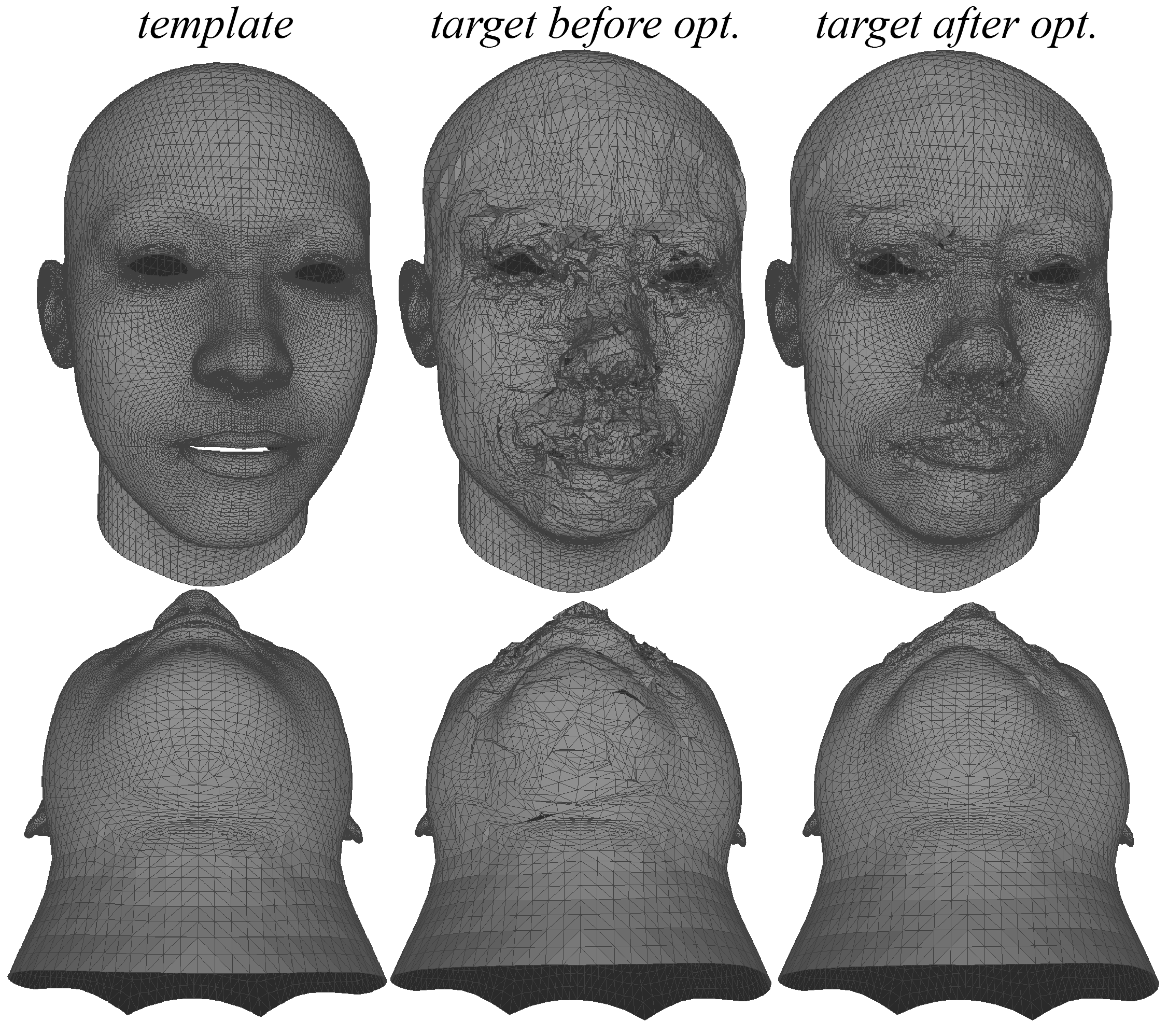}
	\end{center}
	\caption{A failure case for dense registration where the initialization fails to represent the ground-truth surface. Both the near-frontal view and the bottom-up view are shown. Please also zoom in to see the corrective mesh grid routine as a positive effect of the proposed method.}
	\label{fig17}
\end{figure}

The proposed method has been tested with a variety of facial data and can generally ``repair'' the non-uniform mesh grid routine on corresponded faces, even if the initialization is corrupted by tremendous noise (see Fig.~\ref{fig12}) in the tangential direction. However, the proposed method cannot achieve a ``clean'' result when the facial surface is not reconstructed properly by a scanning device. A failure example in the FaceScape dataset is shown in Fig.~\ref{fig17}, where the initial corresponded result by the E-NICP method fails to represent the ground-truth facial surface in the normal direction. The proposed method is not able to solve this problem thoroughly, although most of the non-uniform mesh grid routines are rectified. Therefore, we suggest that the proposed method can expand its applications in combination with the state-of-the-art surface denoising methods.

\subsection{Application to Other Format of Data}
\begin{figure}[htbp]
	\begin{center}
		\includegraphics[width=1\linewidth]{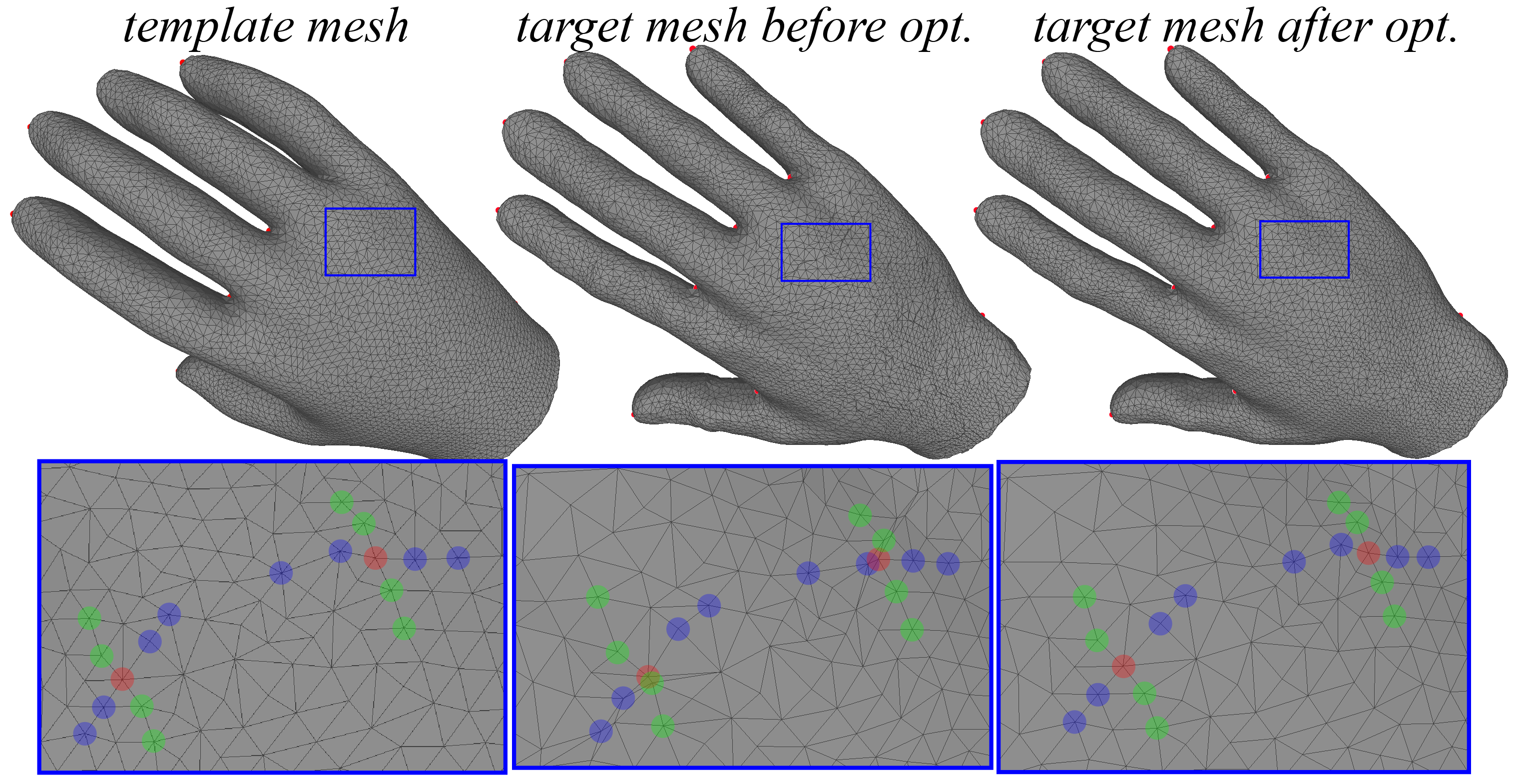}
	\end{center}
	\caption{An example for the optimization of registered hand mesh data. The fixed vertices are shown in red dot.}
	\label{fig18}
\end{figure}

The proposed dividing and diffusing method can be also applied to other format of data which are not limited to face. In fact, it is able to achieve fine-grained correspondence and repair the non-uniform mesh grid routine given a pair of initially registered template and target surface. Fig.~\ref{fig18} shows an example for hand mesh data. The common ground between the face and the hand is that they have both fixed vertices (feature points) and free vertices in the setting for dense registration. The proposed method is general under this assumption, and is thus very useful in a variety of applications for surface registration.

\section{Conclusion}
We propose a dividing and diffusing method for the registration of 3D facial surface in this paper. The proposed method is motivated by proportional segmentation of a line, and alternates between local dividing and global diffusing, to finally achieve pairwise and fine-grained dense correspondence of 3D facial surface. We further elucidate the physical meaning of the proposed method as smooth rearrangement of a local scaling metric for 3D facial shape. Extensive experiments have demonstrated its effectiveness, including computational efficiency, robustness to initializations, metric embedding property, and representation ability of resulted face model. The proposed method can be also used to establish correspondences for other format of surface data, such as hands. Generally, it gives a plausible definition for vertex correspondence on smooth regions for 3D face, and leads to locally coherent details and elegant mesh grid routines for dense registrations, which we hope is helpful for a variety of applications.


There is a remaining issue in the proposed method as well as in other existing works. We link the proposed method to the fundamental forms in Eq.~\ref{e1} for shape analysis in this paper. We assume that the second fundamental form for curvature can be well established by a number of feature points, based on which we optimize the first fundamental form for surface parameterization. However, we observe that some high-curvature features on facial surface cannot be uniquely determined if the number of landmarks is not enough. On the contrary, the accuracy of pre-annotated landmarks is a burden if we include too many of them which are not very salient, especially for raw scanning data with tremendous noise. Therefore, balancing the number and accuracy of facial landmarks as well as including soft constraints versus hard feature constraints deserves further study. In the future, we will explore automatic and accurate landmark localization methods on point clouds for 3D facial data. We will also study ``soft'' method to take into consideration noisy feature matchings for robust 3D face dense correspondence with lifted performance.


%





\ifCLASSOPTIONcaptionsoff
  \newpage
\fi



%
%
%
\bibliographystyle{ieeetr}
\bibliography{egbib}
%

\begin{IEEEbiography}[{\includegraphics[width=1in,height=1.25in,clip,keepaspectratio]{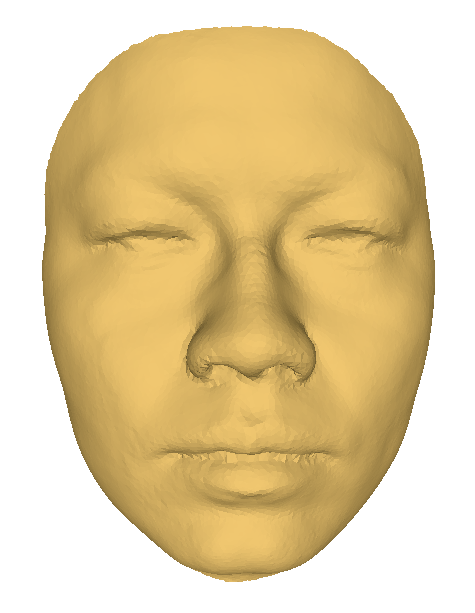}}]{Zhenfeng Fan}
received the B.S. degree in Dalian University of Technology in 2012, M.E. degree in the Institute of Electronic, Chinese Academy of Sciences (CAS) in 2016, and Ph.D. degree in the Institute of Automation, CAS, in 2020. He is currently an assistant professor in the Institute of Computing Technology, CAS. His research interest includes 3D facial analysis, point cloud registration, and image super-resolution.
\end{IEEEbiography}

%

\begin{IEEEbiographynophoto}{Silong Peng}
received the B.S. degree in mathematics from the Anhui University in 1993, and the M.S. and Ph.D. degrees in mathematics from Institute of Mathematics, Chinese Academy of Sciences (CAS), in 1995 and 1998, respectively. From 1998 to 2000, he worked as a postdoctoral researcher in the Institute of Automation, CAS. During this period, he was also a visiting scholar with Department of Mechanics and Mathematics, Lomonosov Moscow State University, Russia. In 2000, he became a full professor of signal processing and pattern recognition in Institute of Automation, CAS. His research interests include wavelets, multi-rate signal processing, and digital image processing.
\end{IEEEbiographynophoto}


\begin{IEEEbiographynophoto}{Shihong Xia}
received the B.S. degree in mathematics from the Sichuan Normal University in 1996, and the M.S. and Ph.D. degree in computer science from the University of Chinese Academy of Sciences (CAS) in 1999 and 2002, respectively. From 2007 to 2008, he was a visiting scholar with the Robotics Institute, School of Computer Science, Carnegie Mellon University, USA. He is currently a professor of the Institute of Computing Technology, CAS, and is the director of the human motion laboratory. His primary research is in the area of computer graphics, virtual reality and artificial intelligence.
\end{IEEEbiographynophoto}




\end{document}